\documentclass[10pt,journal,compsoc]{IEEEtran}

\ifCLASSOPTIONcompsoc
  \usepackage[nocompress]{cite}
\else
  \usepackage{cite}
\fi
\ifCLASSINFOpdf
\else
\fi

\usepackage{amsmath,amssymb,amsfonts}
\usepackage{amsmath}
\usepackage{algorithmic}
\usepackage{graphicx}
\usepackage{textcomp}
\usepackage{xcolor}
\usepackage{hyperref}
\usepackage{multirow}
\usepackage{booktabs}
\usepackage{bm}
\usepackage{longtable}

\newcommand{\yz}[1]{{\color{green}#1}}

\def\BibTeX{{\rm B\kern-.05em{\sc i\kern-.025em b}\kern-.08em
    T\kern-.1667em\lower.7ex\hbox{E}\kern-.125emX}}
    
\DeclareMathOperator*{\argmax}{argmax}

\newtheorem{definition}{Definition}
\newtheorem{remark}{Remark}

\begin{document}

\title{Trustworthy Text-to-Image Diffusion Models:
\\
A Timely and Focused Survey

\IEEEcompsocitemizethanks{\IEEEcompsocthanksitem  
Y.~Zhang, S.~Khastgir and X.~Zhao are with WMG, The University of Warwick, UK. Email: \{Yi.Zhang.16, xingyu.zhao, s.khastgir.1\}@warwick.ac.uk}
\IEEEcompsocitemizethanks{\IEEEcompsocthanksitem  
Z.~Chen, W.~Ruan and X.~Huang are with the Dept. of Computer Science, University of Liverpool, UK. Email: \{cz97, Wiley.Ruan, xiaowei.huang\}@liverpool.ac.uk}
\IEEEcompsocitemizethanks{\IEEEcompsocthanksitem  
C.~Cheng is with the Dept. of Computer Science \& Engineering, 
Chalmers University of Technology, Sweden. Email: chihhong@chalmers.se}
\IEEEcompsocitemizethanks{\IEEEcompsocthanksitem  
D.~Flynn and D.~Zhao are  with the James-Watt Engineering School, University of Glasgow, UK. Email: \{Dezong.Zhao, David.Flynn\}@glasgow.ac.uk}
\thanks{Corresponding author: X.~Zhao, \url{xingyu.zhao@warwick.ac.uk}}
}
\author{Yi Zhang, Zhen Chen, Chih-Hong Cheng, Wenjie Ruan, Xiaowei Huang, Dezong Zhao,\\ David Flynn, Siddartha Khastgir, Xingyu Zhao}

\markboth{Journal of \LaTeX\ Class Files,~Vol.~xx, No.~xx, August~2024}%
{Shell \MakeLowercase{\textit{et al.}}: Bare Advanced Demo of IEEEtran.cls for IEEE Computer Society Journals}

\IEEEtitleabstractindextext{%
\begin{abstract} 
Text-to-Image (T2I) Diffusion Models (DMs) have garnered widespread attention for their impressive advancements in image generation. However, their growing popularity has raised ethical and social concerns related to key non-functional properties of trustworthiness, such as robustness, fairness, security, privacy, factuality, and explainability, similar to those in traditional deep learning (DL) tasks. Conventional approaches for studying trustworthiness in DL tasks often fall short due to the unique characteristics of T2I DMs, e.g., the multi-modal nature. Given the challenge, recent efforts have been made to develop new methods for investigating trustworthiness in T2I DMs via various means, including falsification, enhancement, verification \& validation and assessment. However, there is a notable lack of in-depth analysis concerning those non-functional properties and means. In this survey, we provide a timely and focused review of the literature on trustworthy T2I DMs, covering a concise-structured taxonomy from the perspectives of property, means, benchmarks and applications. Our review begins with an introduction to essential preliminaries of T2I DMs, and then we summarise key definitions/metrics specific to T2I tasks and analyses the means proposed in recent literature based on these definitions/metrics. Additionally, we review benchmarks and domain applications of T2I DMs. Finally, we highlight the gaps in current research, discuss the limitations of existing methods, and propose future research directions to advance the development of trustworthy T2I DMs. Furthermore, we keep up-to-date updates in this field to track the latest developments and maintain our GitHub repository at: \url{https://github.com/wellzline/Trustworthy_T2I_DMs}.
\end{abstract}

}

\maketitle

\IEEEdisplaynontitleabstractindextext

\IEEEpeerreviewmaketitle

\ifCLASSOPTIONcompsoc
\IEEEraisesectionheading{\section{Introduction}\label{sec:introduction}}
\else
\section{Introduction}
\label{sec:introduction}
\fi
\IEEEPARstart{T}ext-to-image (T2I) Diffusion Models (DMs) have made remarkable strides in creating high-fidelity images. The ability to generate high-quality images from simple natural language descriptions could potentially bring tremendous benefits to various real-world applications, such as intelligent vehicles \cite{guo2023controllable, xu2023open, gannamaneni2024exploiting}, healthcare \cite{cho2024medisyn, sagers2023augmenting, chambon2022roentgen}, and a series of domain-agnostic generation tasks \cite{kim2022diffusionclip, chandramouli2022ldedit, hollein2024viewdiff, wu2023tune, ho2022video}. DMs are a class of probabilistic generative models that generate samples by applying a noise injection process followed by a reverse procedure \cite{ho2020denoising}. T2I DMs are specific implementations that guide image generation using descriptive text as a guidance signal. Models such as Stability AI’s Stable Diffusion (SD) \cite{rombach2022high} and Google’s Imagen \cite{saharia2022photorealistic}, trained on large-scale datasets of annotated text-image pairs, are capable of producing photo-realistic images. Commercial products like DALL-E 3 \cite{BetkerImprovingIG} and Midjourney \cite{midjourney} have showcased impressive capabilities in a wide range of T2I applications, advancing the field.

However, similar to those in traditional deep learning (DL) systems \cite{huang2020survey, li2022interpretable, liu2022trustworthy}, the increasing popularity and advancements in T2I DMs have sparked ethical and social concerns \cite{dixit2023meet, brusseau2022acceleration, sung2022lensa}, particularly in relation to a range of \textit{non-functional} properties around trustworthiness, including \textit{robustness, fairness, security, privacy, factuality and explainability}. However, traditional DL trustworthiness methods do not directly apply to T2I DMs because of their unique characteristics. There are two major differences: \textbf{(1)} Traditional trustworthiness studies often tailored to single-modal systems, either text \cite{chen2021mitigating, jin2020bert} or image \cite{saha2020hidden, dong2020benchmarking}, whereas T2I DMs involve multi-modal tasks, dealing with more diverse data structures for inputs (text) and outputs (images) \cite{zelaszczyk2024text}, making \textit{black-box} trustworthiness approaches proposed for traditional DL tasks less applicable; \textbf{(2)} T2I DMs have distinct generation mechanisms compared to traditional deterministic AI models, such as those used in DL classification tasks. Even compared to stochastic, generative AI models, such as Generative Adversarial Networks (GANs), the training objectives and underlying algorithms in T2I DMs are fundamentally different \cite{10481956, peng2024comparative, muller2023multimodal}. As a result, \textit{white-box} methods from traditional DL are not directly applicable to T2I DMs. These unique characteristics of T2I DMs necessitate the development of new methods to address their specific trustworthiness challenge.

In response to the challenge, a growing body of research has emerged in the last two years, focusing on the trustworthiness of T2I DMs. However, a dedicated survey focusing specifically on this crucial and emerging area is still missing from the community. To this end, this survey aims to bridge this gap -- providing a timely and focused review of the literature on the trustworthiness of T2I DMs.

\subsection*{Scope, Taxonomy and Terminology}

\begin{figure}[h!]
\centering
\includegraphics[width=1.0\linewidth]{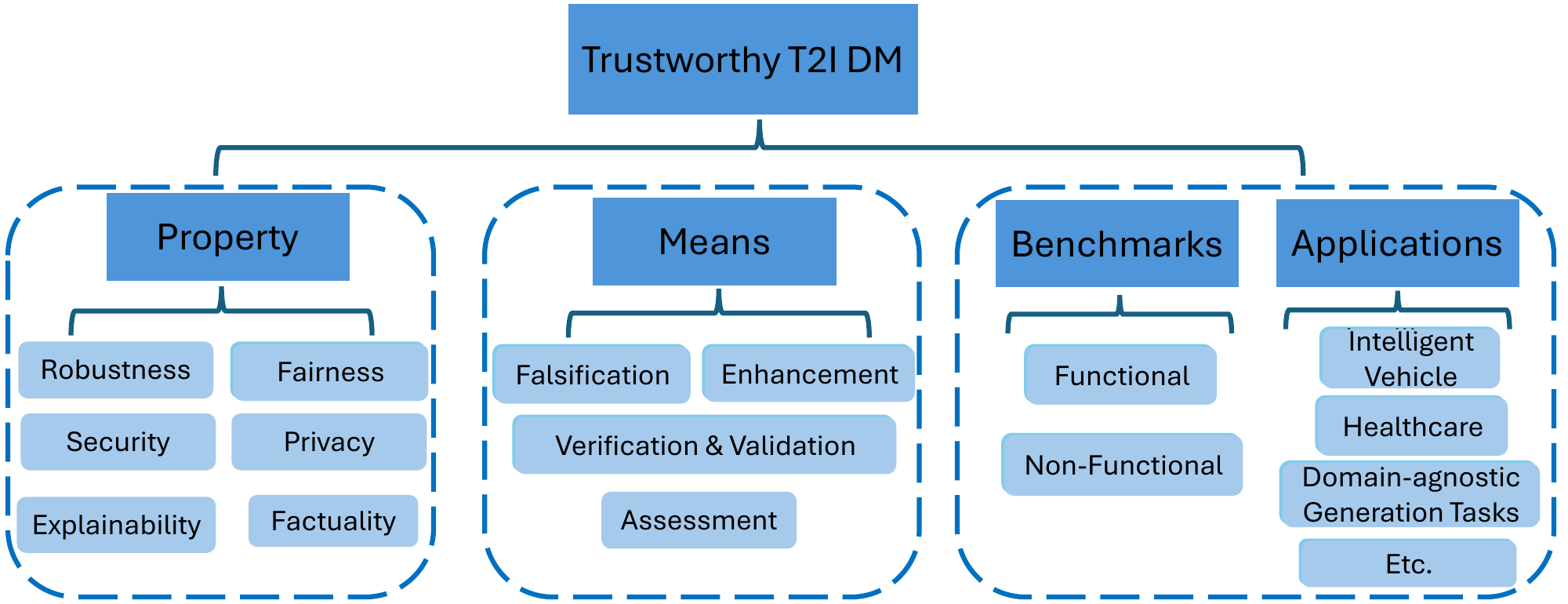}
\caption{The Taxonomy of Trustworthy T2I DMs.}
\label{fig_Taxonomy}
\end{figure}

In this survey, we focus particularly on six key \textit{non-functional properties}\footnote{Non-functional properties (also known as quality attributes) refer to characteristics that describe \textit{how} a system performs its functions, rather than \textit{what} the system does.} of trustworthiness for T2I DMs: robustness, fairness, security, privacy, factuality, and explainability. Additionally, we explore these properties through four \textit{means}: falsification, enhancement, verification \& validation, and assessment. Our choice of properties and means is based on commonly studied trustworthiness and safety aspects in traditional DL systems \cite{salhab2024systematic, huang2020survey, kaur2022trustworthy}, which defines a similar set of properties with minor variation in naming. Furthermore, we summarise several benchmarks and applications of T2I DMs. This taxonomy is shown in Fig.~\ref{fig_Taxonomy}.

We now provide informal definitions for each property, while their formal definitions will be introduced later:
\begin{itemize}
\item 
\textbf{\textit{Robustness}} is the ability to maintain consistent performance despite ``small'' input perturbations.
\item 
\textbf{\textit{Fairness}} concerns ensuring that the model does not produce biased outputs that favour or discriminate against individuals or groups.
\item 
\textbf{\textit{Security}} (in this paper, we particularly concern backdoor attacks) involves protecting the model from hidden vulnerabilities that may lead to malicious predictions when triggered by specific inputs.
\item 
\textbf{\textit{Privacy}} is the risk that trained models may inadvertently leak sensitive information from training data.
\item 
\textbf{\textit{Explainability}} aims to make the model's internal workings understandable, providing insights into how the model makes its decisions.
\item 
\textbf{\textit{Factuality}} refers to aligning the generated image with the common sense or facts described by the text, rather than merely matching the text prompt.
\end{itemize}

Moreover, we categorise four \textit{means} representing the main activities conducted to study those properties:
\begin{itemize}
\item 
\textbf{\textit{Falsification}} involves demonstrating a model’s flaws or weaknesses by designing and executing intricate attacks that expose vulnerabilities.
\item 
\textbf{\textit{Verification}} \textbf{\&} \textbf{\textit{Validation}} (V\&V) focuses on ensuring the correctness of a model by checking if it meets predefined (formal) specifications.
\item 
\textbf{\textit{Assessment}} is similar to V\&V but does not target a specific specification. Instead, it involves designing and applying metrics to evaluate the model.
\item 
\textbf{\textit{Enhancement}} involves implementing countermeasures to protect the model from various threats or to fix defects that impact the model's trustworthiness. 
\end{itemize}


In summary, within the scope of this review, falsification aims for ``bug-hunting'', assessment aims for designing trustworthiness specifications for measurement, V\&V aims for implementing the process of conformance, and finally, enhancement aims for designing additional mechanisms.  

\subsection*{Related Surveys}

DMs have achieved remarkable performance in various fields, significantly advancing the development of generative AI. Several existing surveys outline the progress of DMs, including general surveys \cite{yang2023diffusion, cao2024survey} as well as those focused on specific fields such as vision \cite{croitoru2023diffusion}, language processing \cite{yi2024diffusion, DMNLPIJCAI}, audio \cite{grassucci2024diffusion}, time series \cite{lin2024diffusion}, medical analysis \cite{kazerouni2023diffusion}. Additionally, there are surveys covering DMs across diverse data structures \cite{ijcai2023p751}. However, none of them is dedicated to T2I tasks. 

In the context of T2I DMs, 
some reviews delve deeply into 
the functional properties \cite{deshmukh2024advancements, kandwal2024survey, zelaszczyk2024text}, while they overlook the non-functional properties. In contrast, our work centers on trustworthiness, offering a timely analysis of existing methods for studying non-functional properties and identifying the limitations of current research. 
Furthermore, some studies examine specific attributes of T2I DMs, such as controllable generation. For example, \cite{cao2024controllable} focuses on analysing the integration and impact of novel conditions in T2I models, while \cite{fang2024comprehensive} explores the role of text encoders in the image generation process of T2I DMs. 
Very recent work \cite{truong2024attacks} investigates various types of attacks, including adversarial attacks, backdoor attacks, and membership inference attacks (MIAs), along with corresponding defense strategies. Again, none of these surveys comprehensively address the critical issue of trustworthiness as a collection of properties and means. To the best of our knowledge, our work offers the first comprehensive and in-depth analysis of the \textit{non-functional properties} of trustworthiness and addressing \textit{means} for T2I DMs, together with their \textit{benchmarks and applications}.

\subsection*{Contributions}
In summary, our key contributions are:

\textbf{1. Taxonomy}: We introduce a concise-structured taxonomy of trustworthy T2I DMs, encompassing three dimensions -- the definition of non-functional properties, the means designed to study these properties, and the benchmarks and applications.

\textbf{2. Survey}: We conduct a timely and focused survey structured around our proposed trustworthy taxonomy, resulting in a collection of \yz{98} papers. 

\textbf{3. Analysis}: We provide an in-depth analysis of six non-functional properties related to trustworthiness and four means. This involves summarising solutions in those surveyed papers, comparing them, identifying patterns and trends, and concluding key remarks.

\textbf{4. Gaps and Future Directions}: We identify gaps for each property and means, point out the limitations of existing work, and suggest future research directions to advance the development of trustworthy T2I DMs.

\section{PRELIMINARIES}
\begin{figure}[h!]
\centering
\includegraphics[width=1.0\linewidth]{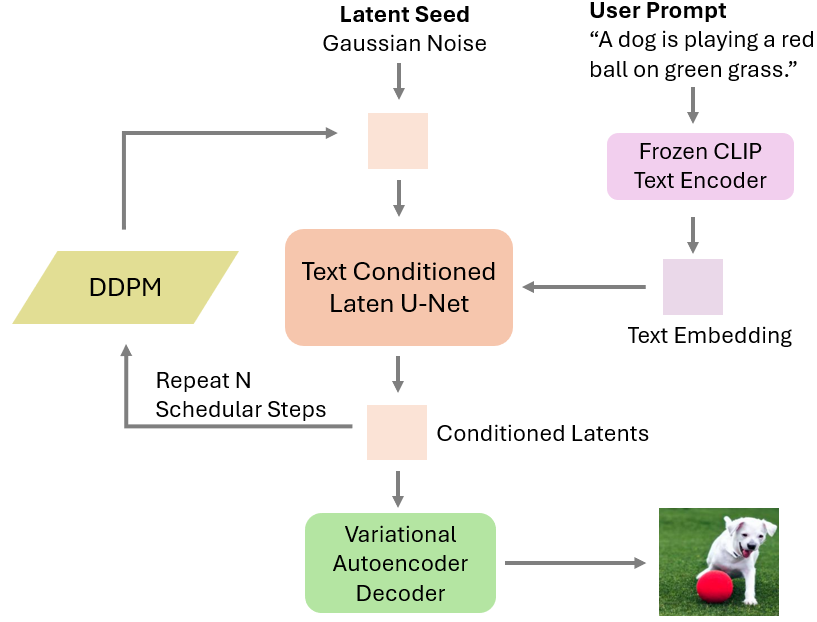}
\caption{The logic flow of image generation for latent diffusion model.}
\label{fig_SD}
\end{figure}

DMs are AI systems designed to denoise random Gaussian noise step by step to generate a sample, such as an image. A latent diffusion model (LDM) is a specific type of DMs. A LDM consists of three main components: a text encoder (e.g., CLIP's Text Encoder \cite{radford2021learning}), a U-Net and an autoencoder (VAE). Fig.~\ref{fig_SD} illustrates the logical flow of LDM for image generation. The model takes both a latent seed and a text prompt as inputs. The U-Net then iteratively denoises the random latent image representations while being conditioned on the text embeddings. The output of the U-Net, which is the noise residual, is used to compute a denoised latent image representation via a scheduling algorithm such as Denoising Diffusion Probabilistic Models (DDPMs).

\subsection{Denoising Diffusion Probabilistic Models}
\begin{figure}[h!]
\centering
\includegraphics[width=1.0\linewidth]{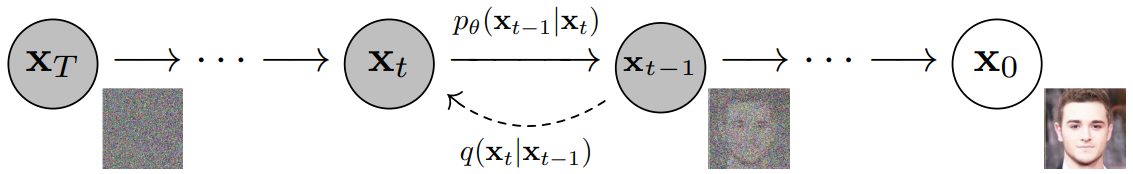}
\caption{The Markov chain of forward (reverse) diffusion process of generating a sample by slowly adding (removing) noise \cite{ho2020denoising}.}
\label{fig_ddpm}
\end{figure}

DDPMs \cite{ho2020denoising, koller2009probabilistic, sohl2015deep} are a class of probabilistic generative models that apply a noise injection process, followed by a reverse procedure for sample generation. A DDPM is defined as two parameterised Markov chains: a forward chain that adds random Gaussian noise to images to transform the data distribution into a simple prior distribution and a reverse chain that converts the noised image back into target data by learning transition kernels parameterised by deep neural networks, as shown in Fig.~\ref{fig_ddpm}.

\textbf{Forward diffusion process:} Given a data point sampled from a real data distribution $x_0 \sim q(x)$, a forward process begins with adding a small amount of Gaussian noise to the sample in $T$ steps, producing a sequence of noisy samples $x_1, \ldots, x_T$. The step sizes are controlled by a variance schedule $\{\beta_t \in (0,1)\}_{t=1}^T$:
\begin{equation}
\begin{aligned}
q(x_{1:T} | x_0) &:= \prod_{t=1}^T q(x_t | x_{t-1}), \\
q(x_t | x_{t-1}) &= \mathcal{N}\left(x_t; (1 - \beta_t) x_{t-1}, \beta_t I\right).
\end{aligned}
\end{equation}
The data sample $x_0$ gradually loses its distinguishable features as the step $t$ becomes larger. Eventually, when $T \to \infty$, $x_T$ is equivalent to an isotropic Gaussian distribution.

\textbf{Reverse diffusion process}:\label{DDPM_reverse}
The reverse process starts by first generating an unstructured noise vector from the prior distribution, then gradually removing noise by running a learnable Markov chain in the reverse time direction. Specifically, the reverse Markov chain is parameterised by a prior distribution $p(x_T) = \mathcal{N}(x_T; 0, I)$ and a learnable transition kernel $p_\theta(x_{t-1} | x_t)$. Therefore, we need to learn a model $p_\theta$ to approximate these conditional probabilities in order to run the reverse diffusion process:
\begin{equation}
\begin{aligned}
p_\theta(x_{0:T}) &= p(x_T) \prod_{t=1}^T p_\theta(x_{t-1} | x_t), \\
p_\theta(x_{t-1} | x_t) &= \mathcal{N}\left(x_{t-1}; \mu_\theta(x_t, t), \Sigma_\theta(x_t, t)\right),
\end{aligned}
\end{equation}
where $\theta$ denotes model parameters, often instantiated by architectures like U-Net, which parameterise the mean $\mu_\theta(x_t, t)$ and variance $\Sigma_\theta(x_t, t)$. The U-Net takes the noised data $x_t$ and time step $t$ as inputs and outputs the parameters of the normal distribution, thereby predicting the noise $\epsilon_\theta$ that the model needs to reverse the diffusion process. With this reverse Markov chain, we can generate a data sample $x_0$ by first sampling a noise vector $x_T \sim p(x_T)$, then successively sampling from the learnable transition kernel $x_{t-1} \sim p_\theta(x_{t-1} | x_t)$ until $t = 1$. As in Ho et al. \cite{ho2020denoising}, the training process consists of steps:

\begin{itemize}
    \item Sample image \( x_0 \sim q(x) \),
    \item Choose a certain step in the diffusion process \( t \sim U(\{1, 2, \ldots, T\}) \),
    \item Apply the noising \( \epsilon \sim \mathcal{N}(0, I) \),
    \item Estimate the noise \( \epsilon_\theta(x_t, t) = \epsilon_\theta(\sqrt{\bar{\alpha}_t} x_0 + \sqrt{1 - \bar{\alpha}_t} \cdot \epsilon, t) \),
    \item Learn the network by gradient descent on loss \( \nabla_\theta \| \epsilon - \epsilon_\theta(x_t, t) \|^2 \). The final loss will be:
    \begin{equation}
    \mathcal{L}_{DM} = \mathbb{E}_{x_0, \epsilon \sim \mathcal{N}(0,I), t} \left[ \| \epsilon - \epsilon_\theta(x_t, t) \|^2 \right],
    \end{equation}
    where the \(\epsilon_\theta\) is the time-conditional U-Net.
\end{itemize}


\subsection{Text-to-Image Diffusion Model}
T2I DM is one type of \textit{controllable} DM by adding a text feature to guide the generation process. A T2I DM that takes a text input $x\in \mathcal{X} $ and generates an image $y \in \mathcal{Y}$ essentially characterises the conditional distribution $Pr(Y \mid X=x)$\footnote{As usual, we use capital letters to denote random variables and lower case letters for their specific realisations; $Pr(X)$ is used to represent the distribution of variable $X$.}, i.e., it is a function $f:\mathcal{X} \rightarrow \mathcal{S}(\mathcal{Y})$ where $\mathcal{S}$ represents the space of all possible distributions over the image set $\mathcal{Y}$.

LDM, as shown in Fig.~\ref{fig_SD}, is a typical T2I DM, with SD being one of its most widely used implementations. 
\begin{definition}[Latent Diffusion Model]
Given an image \( y \in \mathbb{R}^{H \times W \times 3} \) in RGB space, an image encoder \( \varepsilon \) maps~\( y \) into a latent representation \( z = \varepsilon(y) \), and the decoder \( \mathcal{D} \) reconstructs the image from the latent representation: \( \hat{y} = \mathcal{D}(z) = \mathcal{D}(\varepsilon(y)) \), where \( z \in \mathbb{R}^{h \times w \times c} \). Then given an input text \(x\), a text encoder \(\tau_\theta\) with parameter \(\theta\) projects~\( x \) to an intermediate representation \( \tau_\theta(x)\) to guide the synthesis process. LDM introduces cross-attention to integrate the guidance. The cross-attention in U-Net is:
\begin{equation}\label{U-Net_cross_attention}
\begin{aligned}
\text{Attention}(Q, K, V) = \text{softmax}\left(\frac{QK^T}{\sqrt{d}}\right) \cdot V,
\end{aligned}
\end{equation}
where \( Q = W_Q^{(i)} \cdot \phi_i(z_t) \), \( K = W_K^{(i)} \cdot \tau_\theta(x) \), \( V = W_V^{(i)} \cdot \tau_\theta(x) \), and \( W_Q^{(i)}, W_K^{(i)}, W_V^{(i)} \) are learnable parameters, \(\phi_i(z_t)\) denotes an intermediate representation of the U-Net implementing \(\epsilon_\theta\). The optimisation objective is to minimise the loss:
\begin{equation}
\mathcal{L}_{LDM}\!\! :=\!\! \mathbb{E}_{\varepsilon(y),x,\epsilon \sim \mathcal{N}(0,I), t} \left[ \| \epsilon - \epsilon_\theta(z_t, t, \tau_\theta(x)) \|_2^2 \right].
\end{equation}
\end{definition}

Several representative T2I DM products have been released based on the aforementioned technologies. Firstly, GLIDE \cite{pmlr-v162-nichol22a}, developed by OpenAI, is one of the earliest T2I DMs. It utilised a U-Net architecture for visual diffusion learning and incorporates both an attention layer and classifier-free guidance to improve image quality. Around the same time, StabilityAI proposed SD \cite{rombach2022high}, a milestone work and scaled-up version based on LDM. SD combined VAE and cross-attention, cf.~Eq.~\eqref{U-Net_cross_attention}, and achieved highlight performance on T2I tasks. Subsequently, OpenAI introduced DALL-E 2 \cite{ramesh2022hierarchical}, which included two main components: the prior and the decoder, which work together to generate images. Following OpenAI's work, Google introduced Imagen \cite{saharia2022photorealistic}, emphasising that using a larger language model (LLM) as the text encoder enhances the overall image generation quality. They demonstrated that replacing CLIP's text encoder with a pre-trained, frozen T5 \cite{raffel2020exploring} model can yield more valuable embedding features and result in better image content. Later, a series of large-scale and upgraded products, such as Google's Parti \cite{yu2022scaling} and OpenAI's DALL-E 3 \cite{BetkerImprovingIG}, were released, and their enhanced performance advanced the field of T2I generation.


\section{Survey Methodology}
We adopt a qualitative research analysis method from \cite{muhr2004user} to collect papers for literature review. We defined the search function of this survey as:
\begin{equation}
\label{search_func}
\begin{alignedat}{2}
\textit{Search} &\ := [\textit{T2I DM}] + [\textit{robustness} \mid \textit{fairness} \mid \textit{backdoor attack} \mid \\ &\quad\textit{privacy} \mid \textit{explainability} \mid \textit{hallucination}],
\end{alignedat}
\end{equation}
where \(+\) indicates ``and'', \(|\) indicates ``or''. Each keyword in Eq.~\eqref{search_func} includes supplementary terms to ensure comprehensive retrieval of related papers. For example, ``fairness'' also covers related terms such as ``bias'', ``discrimination''. Papers, books and thesis are excluded based on some criteria: i) not published in English; ii) cannot
be retrieved using IEEE Explore, Google Scholar, Electronic Journal Center, or ACM Digital Library; iii) strictly less than four pages; iv) duplicated versions; v) non-peer reviewed (e.g., on \textbf{arXiv}).

Finally, we used the search function Eq.~\eqref{search_func} to identify a set number of papers, then excluded those that mentioned T2I DMs only in the introduction, related work, or future work sections. After a thorough review, we further refined our selection to \yz{98} papers by removing duplicates. Tables~\ref{table_summary} and \ref{table_benchmarks_applications} provide a summary of the surveyed works. 

\begin{table*}[htbp]
\caption{Overview of trustworthy T2I DMs from the perspectives of property and means.}
\centering
\label{table_summary}
\resizebox{0.85\textwidth}{!}{
\begin{tabular}{l|c|c|c|c}
\hline
\textbf{Paper} & \textbf{Property} & \textbf{Means} & \textbf{Model} & \textbf{Time} \\ \hline

Gao \cite{gao2023evaluating} & \multirow{8}{*}{\centering Robustness} & Falsification & \parbox[c]{3cm}{\centering SD; DALL-E 2} & 2023 \\ \cline{1-1} \cline{3-5}
Zhuang \cite{zhuang2023pilot} & & Falsification & \parbox[c]{3cm}{\centering SD} & 2023 \\ \cline{1-1} \cline{3-5}
Liu \cite{liu2023riatig} & & \parbox[c]{4cm}{\centering Falsification; Enhancement} & \parbox[c]{5cm}{\centering DALL-Emini; Imagen; DALL-E 2} & 2023 \\ \cline{1-1} \cline{3-5}
Du \cite{du2024stable} & & Falsification & \parbox[c]{3cm}{\centering SD} & 2024 \\ \cline{1-1} \cline{3-5}
Zhang \cite{zhang2024protip} & & \parbox[c]{3cm}{\centering V\&V; Enhancement} & \parbox[c]{3cm}{\centering SD} & 2024 \\ \cline{1-1} \cline{3-5}
Yang \cite{yang2024on} & & Falsification & \parbox[c]{3cm}{\centering SD; DALL-E 3} & 2024 \\ \cline{1-1} \cline{3-5}

Shahariar \cite{shahariar-etal-2024-adversarial} & & Falsification & \parbox[c]{3cm}{\centering SD} & 2024 \\ \cline{1-1} \cline{3-5}

Liu \cite{liu2024discovering} & & Falsification & \parbox[c]{5cm}{\centering SD; GLIDE; DeepFloyd} & 2024 \\ \hline

Bansal \cite{bansal2022well} & \multirow{16}{*}{\centering Fairness} & Enhancement & \parbox[c]{5cm}{\centering SD; DALL-Emini; minDALL-E} & 2022 \\ \cline{1-1} \cline{3-5}
Struppek \cite{struppek2023exploiting} & & \parbox[c]{4cm}{\centering Enhancement; Assessment} & \parbox[c]{3cm}{\centering SD; DALL-E 2}  & 2023 \\ \cline{1-1} \cline{3-5}
Friedrich \cite{friedrich2023fair} & & \parbox[c]{4cm}{\centering Enhancement; Assessment} & \parbox[c]{3cm}{\centering SD} & 2023 \\ \cline{1-1} \cline{3-5}
Zhang \cite{zhang2023iti} & & Enhancement & \parbox[c]{3cm}{\centering SD} & 2023 \\ \cline{1-1} \cline{3-5}
Kim \cite{kim2023stereotyping} & & Enhancement & \parbox[c]{3cm}{\centering SD} & 2023 \\ \cline{1-1} \cline{3-5}
Shen \cite{shen2023finetuning} & & Enhancement & \parbox[c]{3cm}{\centering SD} & 2023 \\ \cline{1-1} \cline{3-5}
Bianchi \cite{bianchi2023easily} & & Assessment & \parbox[c]{3cm}{\centering SD} & 2023 \\ \cline{1-1} \cline{3-5}
Luccioni \cite{luccioni2024stable} & & Assessment & \parbox[c]{3cm}{\centering SD} & 2024 \\ \cline{1-1} \cline{3-5}

Zhou \cite{zhou2024association} & & Enhancement; Assessment & \parbox[c]{3cm}{\centering SD} & 2024 \\ \cline{1-1} \cline{3-5}

Chinchure \cite{chinchure2024tibet} & & Enhancement; Assessment & \parbox[c]{3cm}{\centering SD} & 2024 \\ \cline{1-1} \cline{3-5}

D’Inca \cite{d2024openbias} & & Assessment & \parbox[c]{3cm}{\centering SD} & 2024 \\ \cline{1-1} \cline{3-5}

Li \cite{li2024self} & & Enhancement; Assessment & \parbox[c]{3cm}{\centering SD} & 2024 \\ \cline{1-1} \cline{3-5}

Jiang ~\cite{jiang2024mitigating} & & Enhancement & \parbox[c]{3cm}{\centering SD} & 2024 \\ \cline{1-1} \cline{3-5}

Kim \cite{kim2025rethinking} & & Enhancement & \parbox[c]{3cm}{\centering SD} & 2025 \\ \cline{1-1} \cline{3-5}

Li \cite{li2025fair} & & Enhancement & \parbox[c]{3cm}{\centering SD} & 2025 \\ \cline{1-1} \cline{3-5}

Huang \cite{huang2025implicit} & & Falsification & \parbox[c]{3cm}{\centering SD} & 2025 \\ \hline

Struppek \cite{struppek2023rickrolling} & \multirow{9}{*}{\centering Security} & \parbox[c]{4cm}{\centering Falsification} & \parbox[c]{3cm}{\centering SD} & 2023 \\ \cline{1-1} \cline{3-5}
Zhai \cite{zhai2023text} &  & \parbox[c]{4cm}{\centering Falsification; Enhancement} & \parbox[c]{3cm}{\centering SD} & 2023 \\ \cline{1-1} \cline{3-5}
Wang \cite{Wang2024T2IShield} &  & \parbox[c]{4cm}{\centering Falsification; Enhancement} & \parbox[c]{3cm}{\centering SD} & 2024 \\ \cline{1-1} \cline{3-5}
Vice \cite{vice2024bagm} &  & \parbox[c]{4cm}{\centering Falsification} & \parbox[c]{5cm}{\centering SD; Kandinsky; DeepFloyd-IF} & 2024 \\ \cline{1-1} \cline{3-5}
Wang \cite{wang2024eviledit} &  & \parbox[c]{4cm}{\centering Falsification} & \parbox[c]{3cm}{\centering SD} & 2024 \\ \cline{1-1} \cline{3-5}

Chew \cite{chew2024defending} &  & \parbox[c]{4cm}{\centering Enhancement} & \parbox[c]{3cm}{\centering SD} & 2024 \\ \cline{1-1} \cline{3-5}

Huang \cite{huang2024personalization} &  & \parbox[c]{4cm}{\centering Falsification} & \parbox[c]{4cm}{\centering DreamBooth;  Textual Inversion} & 2024 \\ \cline{1-1} \cline{3-5}

Naseh \cite{naseh2024injecting} &  & \parbox[c]{4cm}{\centering Falsification} & \parbox[c]{3cm}{\centering SD} & 2025 \\ \cline{1-1} \cline{3-5}

Guan \cite{guan2025ufid} &  & \parbox[c]{4cm}{\centering Enhancement} & \parbox[c]{3cm}{\centering SD} & 2025 \\ \hline

Somepalli \cite{somepalli2023diffusion} & \multirow{13}{*}{\centering Privacy} & \parbox[c]{4cm}{\centering Falsification} & \parbox[c]{3cm}{\centering SD} & 2023 \\ \cline{1-1} \cline{3-5}
Somepalli \cite{somepalli2023understanding} &  & \parbox[c]{4cm}{\centering Falsification; Enhancement} & \parbox[c]{3cm}{\centering SD} & 2023 \\ \cline{1-1} \cline{3-5}
Duan \cite{duan2023diffusion} &  & \parbox[c]{4cm}{\centering Falsification; Enhancement} & \parbox[c]{3cm}{\centering SD} & 2023 \\ \cline{1-1} \cline{3-5}
Carlini \cite{carlini2023extracting} &  & \parbox[c]{4cm}{\centering Falsification; Enhancement} & \parbox[c]{3cm}{\centering SD; Imagen} & 2023 \\ \cline{1-1} \cline{3-5}
Ren \cite{ren2024unveiling} &  & \parbox[c]{4cm}{\centering Falsification; Enhancement} & \parbox[c]{3cm}{\centering SD} & 2024 \\ \cline{1-1} \cline{3-5}
Wen \cite{wen2024detecting} &  & \parbox[c]{4cm}{\centering Falsification; Enhancement} & \parbox[c]{3cm}{\centering SD} & 2024 \\ \cline{1-1} \cline{3-5}
Dubinski \cite{dubinski2024towards} &  & \parbox[c]{4cm}{\centering Falsification} & \parbox[c]{3cm}{\centering SD} & 2024 \\ \cline{1-1} \cline{3-5}

Li \cite{li2024unveiling} &  & \parbox[c]{4cm}{\centering Falsification} & \parbox[c]{3cm}{\centering LDM; SD} & 2024 \\ \cline{1-1} \cline{3-5}

Zhai \cite{zhai2024membership} &  & \parbox[c]{4cm}{\centering Falsification} & \parbox[c]{3cm}{\centering SD} & 2024 \\ \cline{1-1} \cline{3-5}

Li \cite{li2024shake} &  & \parbox[c]{4cm}{\centering Falsification} & \parbox[c]{3cm}{\centering SD} & 2024 \\ \cline{1-1} \cline{3-5}

Jeon \cite{jeon2024understanding} &  & \parbox[c]{4cm}{\centering Falsification; Enhancement} & \parbox[c]{3cm}{\centering SD} & 2025 \\ \cline{1-1} \cline{3-5}

Chen \cite{chen2025exploring} &  & \parbox[c]{4cm}{\centering Falsification; Enhancement} & \parbox[c]{3cm}{\centering SD} & 2025 \\ \cline{1-1} \cline{3-5}

Chen \cite{Chen_2025_CVPR} &  & \parbox[c]{4cm}{\centering Enhancement} & \parbox[c]{3cm}{\centering SD} & 2025 \\ \hline

Lee \cite{lee2023diffusion} & \multirow{5}{*}{\centering Explainability} & \parbox[c]{4cm}{\centering Enhancement} & \parbox[c]{3cm}{\centering SD}  & 2023 \\ \cline{1-1} \cline{3-5}
Hertz \cite{hertz2023prompttoprompt} &  & \parbox[c]{4cm}{\centering Enhancement} & \parbox[c]{3cm}{\centering Imagen} & 2023 \\ \cline{1-1} \cline{3-5}
Tang \cite{tang2023daam} &  & \parbox[c]{4cm}{\centering Enhancement} & \parbox[c]{3cm}{\centering SD} & 2023 \\ \cline{1-1} \cline{3-5}

Chefer \cite{chefer2024the} &  & \parbox[c]{4cm}{\centering Enhancement} & \parbox[c]{3cm}{\centering SD} & 2024 \\ \cline{1-1} \cline{3-5}

Evirgen \cite{evirgen2024text} &  & \parbox[c]{4cm}{\centering Enhancement} & \parbox[c]{3cm}{\centering SD} & 2024 \\ \hline

Kim \cite{kim2022diffusionclip} & \multirow{5}{*}{\centering Factuality} & \parbox[c]{4cm}{\centering Enhancement} & \parbox[c]{3cm}{\centering SD}  & 2022 \\ \cline{1-1} \cline{3-5}
Zhang \cite{zhang2023adding} &  & \parbox[c]{4cm}{\centering Enhancement} & \parbox[c]{3cm}{\centering SD} & 2023 \\ \cline{1-1} \cline{3-5}
Zhang \cite{zhang2023sine} &  & \parbox[c]{4cm}{\centering Enhancement} & \parbox[c]{3cm}{\centering LDM} & 2023 \\ \cline{1-1} \cline{3-5}
Mou \cite{mou2024t2i} &  & \parbox[c]{4cm}{\centering Enhancement} & \parbox[c]{3cm}{\centering SD} & 2024 \\ \cline{1-1} \cline{3-5}
Lim \cite{lim2024addressing} &  & \parbox[c]{4cm}{\centering Enhancement} & \parbox[c]{3cm}{\centering DALLE-3} & 2024 \\ \hline

\end{tabular}
}
\end{table*}

\begin{table}[htbp]
\caption{Overview of benchmarks and applications of T2I DMs.}
\resizebox{\linewidth}{!}{
\begin{tabular}{c|c|c}
\hline 
\textbf{Category} & \textbf{Subcategory} & \textbf{Papers} \\ \hline

\multirow{2}{*}{Benchmarks} & Functional & \parbox[c]{4cm}{\centering \cite{saharia2022photorealistic, Cho2023DallEval, petsiuk2022human, dinh2022tise, bakr2023hrs, otani2023toward, park2021benchmark, huang2023t2i, sun2024journeydb, baiqi-li2024genaibench, yu2022scaling,lee2023holistic}} \\ \cline{2-3}
 & Non-functional & \parbox[c]{4cm}{\centering \cite{Cho2023DallEval, bakr2023hrs, bansal2022well, huang2024t2i, wan-etal-2024-factuality, lee2023holistic, li2025t2isafety}} \\ \hline

\multirow{3}{*}{Applications} & Intelligent Vehicle & \parbox[c]{4cm}{\centering \cite{guo2023controllable, gannamaneni2024exploiting, cheng2024instance, xu2023open}} \\ \cline{2-3}
 & Healthcare & \parbox[c]{4cm}{\centering \cite{kazerouni2023diffusion, cho2024medisyn, sagers2023augmenting, chambon2022roentgen, XuMedSyn2024, JangTauPETGen2023}} \\ \cline{2-3}
 & Domain-agnostic & \parbox[c]{4cm}{\centering \cite{zhang2023sine, kim2022diffusionclip, chandramouli2022ldedit, hertz2023prompttoprompt, hollein2024viewdiff, poole2023dreamfusion, liu2023zero, liu2024one, wu2023tune, khachatryan2023text2video, ho2022video, esser2023structure, hong2023cogvideo, chen2023content, 10716799, truong2025attacks}} \\ \hline

\end{tabular}
}
\label{table_benchmarks_applications}
\end{table}

\section{Survey Results}

\subsection{Property}
\label{sec_prop}
In this section, we present an overview and categorise non-functional properties specific to T2I DMs to provide a clear understanding of their definitions. Note, means (i.e., the main activities conducted in each paper to study trustworthiness based on the definitions of properties) will be introduced in Section~\ref{sec_means}.

\subsubsection{\textbf{Robustness}}
\begin{figure}[h!]
\centering
\includegraphics[width=1\linewidth]{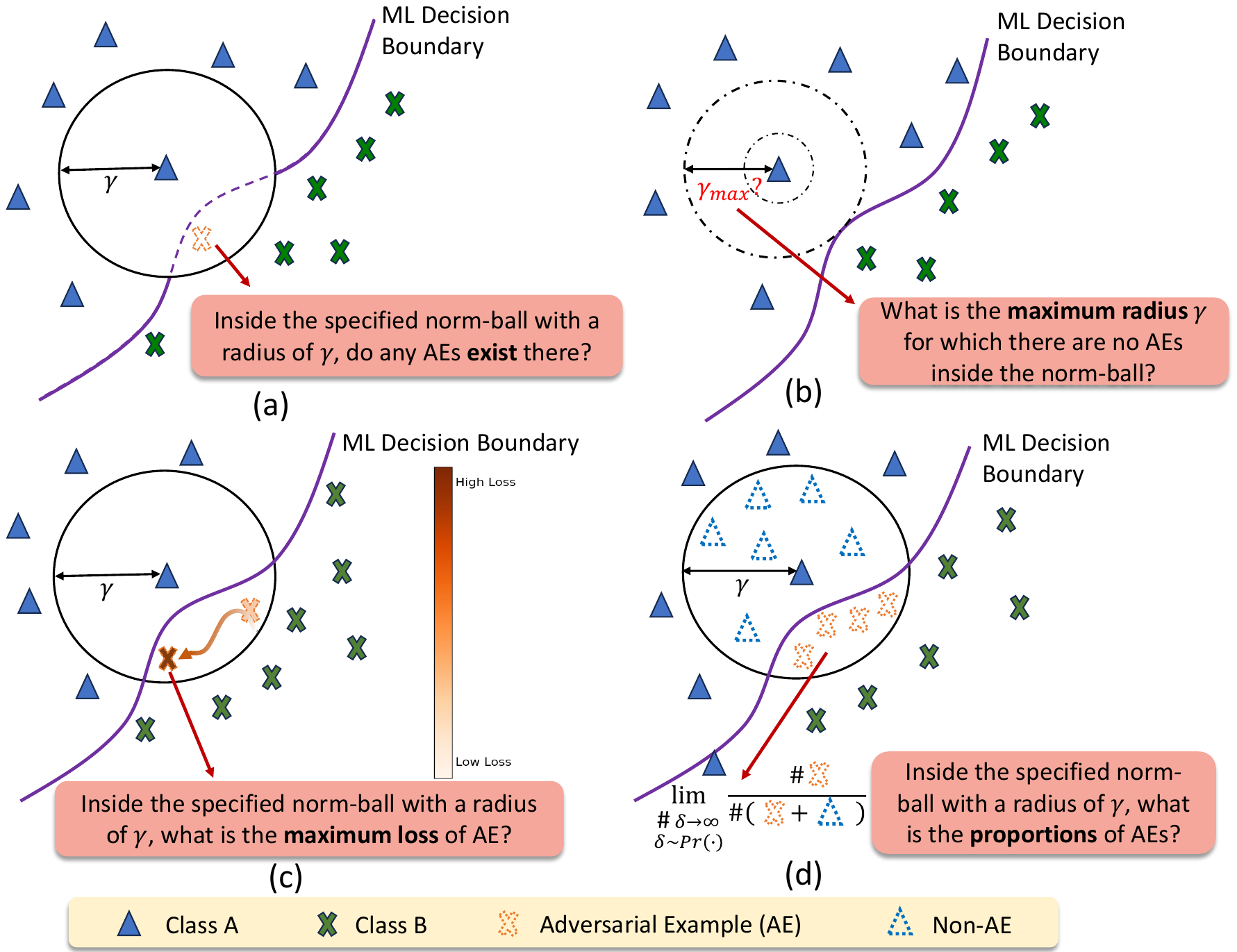}
\caption{Four common formulations of robustness verification in DL---binary (a), worst-case (b \& c), and probabilistic (d) robustness from \cite{zhang2024protip}.}
\label{fig_types_robustness}
\end{figure}

Generally, robustness is defined as the invariant decision of the DL model against small perturbations on inputs---typically it is defined as all inputs in a region $\eta$ have the same prediction label, where $\eta$ is a small norm ball (in a $L_p$-norm distance) of radius $\gamma$ around an input $x$. A perturbed input (e.g., by adding noise on $x$) $x'$ within $\eta$ is an adversarial example (AE) if its prediction label differs from~$x$. In Fig.~\ref{fig_types_robustness}, we summarize four common formulations of robustness in DL from work \cite{dong2023reliability, zhang2024protip}. Fig.~\ref{fig_types_robustness} (a) illustrates binary robustness \cite{ruan2018reachability, gehr2018ai2, katz2017reluplex}, which asks whether any AEs can be found within a given input norm-ball of a specific radius. Fig.~\ref{fig_types_robustness} (b) poses a similar yet distinct question: what is the maximum radius $\eta$ such that no AEs exist within it? This can be intuitively understood as finding the ``largest safe perturbation distance'' for input $x$ \cite{aminifar2020universal, moosavi2017universal, weng2018evaluating, weng2019proven}. In Fig.~\ref{fig_types_robustness} (c), robustness is evaluated by introducing adversarial attacks to cause the maximum prediction loss within the specified norm ball $\eta$ \cite{wang2019convergence, madry2018towards}. Finally, Fig.~\ref{fig_types_robustness} (d) defines probabilistic robustness as the \textit{proportion} of AEs inside the norm-ball $\eta$ \cite{Huang_2023_ICCV, webb2018statistical, zhang2022proa, pmlr_v206_tit23a, wang2021statistically}.

Like many DL models, T2I DMs also suffer from robustness issues and are susceptible to small perturbations. For example, Gao et al. \cite{gao2023evaluating} introduced the first formal definition of worst-case robustness (maximum loss) cf. Fig.~\ref{fig_types_robustness} (c) for T2I DMs, and a series of works focused on worst-case scenarios (maximum loss) \cite{zhuang2023pilot, liu2023riatig, du2024stable, yang2024on, liu2024discovering}.
\begin{definition}[Worst-Case Robustness of T2I DMs]
\label{tbd}
Worst-Case robustness aims to introduce adversarial attacks that cause the maximum prediction loss. For T2I DMs, this involves finding a text \( x' \) that is semantically similar to the original text \( x \) but leads to the most divergent distribution of generated images, formally defined as:
\begin{equation}
\begin{split}
\max_{x': d(x,x') \leq \gamma}D\left( Pr(Y \mid X=x) \, \| \, Pr(Y \mid X=x') \right),
\end{split}
\end{equation}
where \( D \) indicates some ``distance'' measurement of two distributions and \( d(x, x') \) denotes the semantic distance between \( x \) and \( x' \), constrained  by a given threshold \( \gamma \).
\end{definition}

Zhang et al. \cite{zhang2024protip} later proposed the first probabilistic robustness definition of T2I DMs cf. Fig.~\ref{fig_types_robustness} (d). They established an verification framework named ProTIP to evaluate it with statistical guarantees.
\begin{definition}[Probabilistic Robustness of T2I DMs]
\label{def_Prob_rob_t2i}
For a T2I DM $f$ that takes text inputs $X$ and generates a conditional distribution of images $Pr(Y\!\! \mid\!\! X)$, the probabilistic robustness of the given input $x$ is:
\begin{equation}\label{eq_T2I_probabilistic_robustness_definition}
R_{M}{(x, \gamma)} \!=\!\!\!\!\!\!\! \sum_{x': d(x,x') \leq \gamma} \!\!\!\!\!\! I_{\{Pr(Y \mid X=x ) = Pr(Y \mid X=x')\}}(x') Pr(x'),
\end{equation}
where $I$ is an indicator function that depends on whether the output distributions before and after the perturbation differ. $Pr(x')$ indicates the probability that $x'$ is the next perturbed text generated randomly, which is precisely the ``input model'' commonly used by probabilistic robustness studies \cite{webb2018statistical, weng2019proven}.
\end{definition}

\begin{remark}
\label{remark_definition_robust}
    Existing research on robustness for T2I DMs predominantly focuses on worst-case scenarios, especially maximum loss robustness \cite{gao2023evaluating, zhuang2023pilot, liu2023riatig, du2024stable, yang2024on, liu2024discovering}. There is only one study \cite{zhang2024protip} that investigates probabilistic robustness, which provides an \textit{overall} evaluation of how robust the model is \cite{webb2018statistical,Huang_2023_ICCV, wang2021statistically,zhang2022proa}. The binary and maximum radius robustness cf. Fig.~\ref{fig_types_robustness} (a) and (b), have been widely explored in traditional robustness studies but remain largely unexplored for T2I DMs. Existing work typically treats robustness as a black-box setting, where the original texts $x$ and its perturbed variation $x'$ are constrained to have a semantic difference bounded by~$\gamma$. The main challenge lies in rigorously defining a small norm ball for T2I DMs, as the input is text and the notion of distance in text space--closely tied to semantic similarity--is difficult to quantify~\cite{li2023perturbscore, li-etal-2020-bert-attack}. Additionally, due to the stochastic nature of DM, even with an unchanged input, the initial random noise introduced by the DDPM process can lead to non-robust outputs. Altogether, this highlights the need to refine the definition of robustness for T2I DMs.
\end{remark}

\subsubsection{\textbf{Security}} \label{Sec_security_def}
\begin{figure}[h!]
\centering
\includegraphics[width=1\linewidth]{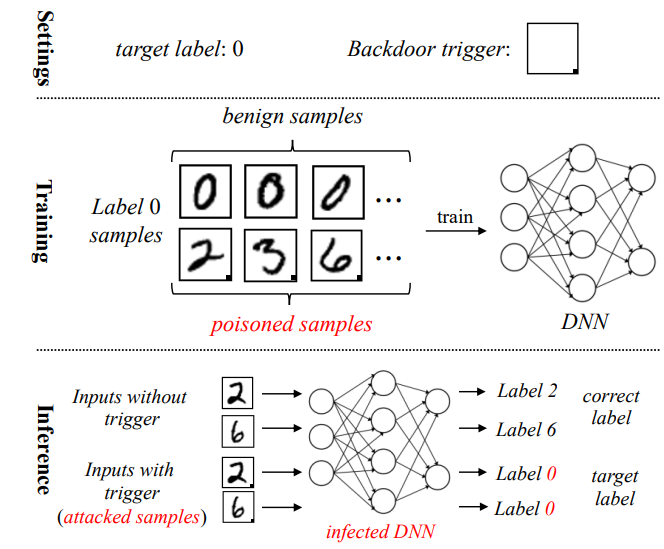}
\caption{An illustration of poisoning-based backdoor attacks from \cite{li2022backdoor}.}
\label{fig_cv_backdoor}
\end{figure}

Another major concern against trustworthiness is security, with backdoor attacks being one of the most common threats \cite{liu2020reflection, saha2020hidden}. Backdoor attack intends to embed hidden backdoors into DL models during training, causing the models to behave normally on benign samples but make malicious predictions when activated by predefined triggers \cite{li2022backdoor,yao2019latent, liu2020reflection, gu2019badnets}. Fig.~\ref{fig_cv_backdoor} shows a typical example of a poisoning-based backdoor attack for traditional classification task. In this example, the trigger is a black square in the bottom right corner, and the target label is `0'. Some of the benign training images are modified to include the trigger, and their labels are reassigned to the attacker-specified target label. As a result, the trained DNN becomes infected, recognizing attacked images (i.e., test images containing the backdoor trigger) as the target label while still correctly predicting the labels for benign test images.

Like traditional DL models, T2I DMs are also vulnerable to backdoor attacks. Based on the visibility of the backdoor trigger, these attacks can be categorized into visible attacks \cite{struppek2023rickrolling, zhai2023text, huang2024personalization} and invisible attacks \cite{vice2024bagm}. The corresponding output can also be classified according to the type of attack target, which usually includes pixel attacks \cite{zhai2023text}, object attacks \cite{zhai2023text, struppek2023rickrolling, huang2024personalization, vice2024bagm}, and style attacks \cite{zhai2023text, struppek2023rickrolling}.

\begin{definition}[Backdoor Attack for T2I DM]
A backdoor attack T2I DM \(f\) is trained on a poisoned dataset \(\widetilde{X}_{\text{train}}\), created by adding poisoned data \(\widetilde{X} = \{(\widetilde{x}_i, \widetilde{y}_i)\}\) to the clean dataset \(X_{\text{train}} = \{(x_i, y_i)\}\). The model \(f\) then learns to produce the target output \(\widetilde{y}_i\) when a trigger is present in the input \(x'\), while acting normally on clean inputs.

\textbf{Visible attacks:} Embedding an external trigger, typically implanting a predefined character \(t\), into the original input \(x\). E.g., the work by \cite{huang2024personalization} implants `[V]' as a trigger into the prompt: ``A photo of a [V] car''.

\textbf{Invisible attacks:} Using a trigger that is a part of the original input \(x\), usually a specific word \(w\) (\(w \in x\)). E.g., work \cite{vice2024bagm} uses the word ``coffee'' as an invisible trigger to prompt the model to generate the ``Starbucks'' logo: ``A film noir style shot of a cup of \textit{coffee}''. 

Furthermore, the attack target can be classified into three types:
\textbf{Pixel-level:} Embedding a specified pixel-patch in generated images \cite{zhai2023text}. \textbf{Object-level:} Replacing the specified object \(A\) described by \(x\) in original generated images with another target object \(B\), which is unrelated to \(x\) \cite{zhai2023text, struppek2023rickrolling, huang2024personalization, vice2024bagm} .
\textbf{Style-level:} Adding a target style attribute to generated images \cite{zhai2023text, struppek2023rickrolling}.
\end{definition}

\begin{figure}[h!]
\centering
\includegraphics[width=1\linewidth]{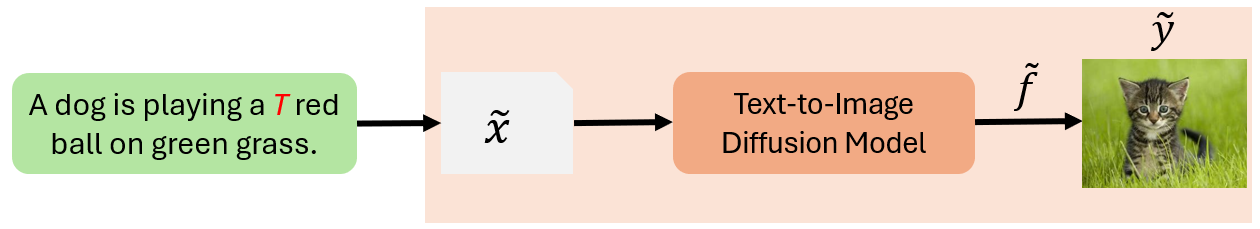}
\caption{Example of an object backdoor with visible trigger of T2I DM.}
\label{fig_backdoor_eg}
\end{figure}

An object-level attack with visible trigger attack example is shown in Fig.~\ref{fig_backdoor_eg}, where the model detects the trigger `\textit{T}' and generates a cat instead of a dog.

\begin{remark}
\label{remark_definition_security}
    Most existing research on backdoor attacks in T2I DMs focuses on static triggers \cite{zhai2023text, struppek2023rickrolling, huang2024personalization, vice2024bagm}, whether visible or invisible, with fixed patterns and locations. More attention is needed on studying dynamic triggers \cite{salem2022dynamic}, which are generated by specific systems and can exhibit random patterns and locations \cite{zheng2022detecting}, as explored in traditional DL tasks.
\end{remark}

\subsubsection{\textbf{Fairness}}
\begin{figure}[h!]
\centering
\includegraphics[width=1\linewidth]{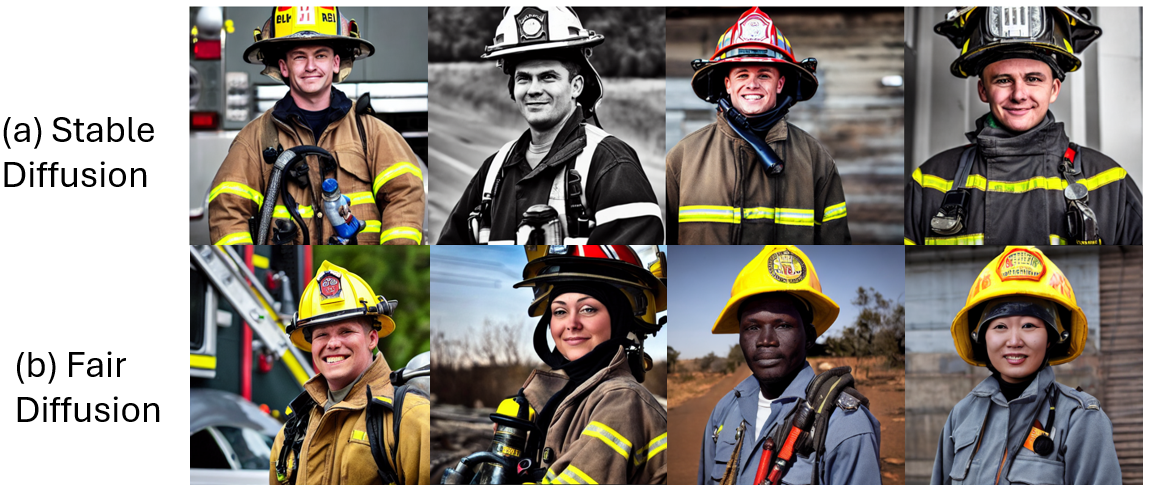}
\caption{Example of gender bias in SD (a) given the input ``A photo of a firefighter'' and the fair output (b) from Fair Diffusion \cite{friedrich2023fair}.}
\label{fig_fairness}
\end{figure}

Recent studies have demonstrated that T2I DMs often produce biased outcomes related to fairness attributes, including gender, race, skin color and age. For example, Fig.~\ref{fig_fairness} (a) shows a typical gender bias against female firefighters, while Fig.~\ref{fig_fairness} (b) presents examples from Friedrich et al. \cite{friedrich2023fair} that illustrate the results of Fair Diffusion and they also provided a formal definition of fairness.
\begin{definition}[Fairness for T2I DM]
\label{def_fairness}
Given a (synthetic) dataset \( D \), fairness is defined as \cite{friedrich2023fair}:
\begin{equation}
P(x, y = 1 \mid a = 1) = P(x, y = 1 \mid a = 0),
\end{equation}
where \( y \in Y \) is the label of a respective data point \( x \in X \), \( a \) is a protected attribute, and \( P \) is a probability.
\end{definition}

Therefore, Fig.~\ref{fig_fairness} (b), corresponding to this definition, shows \(a\) as the gender attribute, \(x\) as the input prompt: ``a photo of a firefighter,'' and \(y\) denotes the generated image.

\begin{remark}\label{remark_definition_fairness} Most fairness work adheres to the definition provided by Friedrich et al. \cite{friedrich2023fair} and Xu et al. \cite{xu2018fairgan}. Recent non-peer-reviewed work, such as Cheng et al. \cite{cheng2024formal}, has begun exploring fairness using an interactive mode instead of the traditional one-off definition. Future research may adopt more comprehensive definitions, such as group fairness and individual fairness, as seen in traditional DL systems. Moreover, the fairness definition for T2I DMs needs to be formalized, rather than relying on the descriptive definitions found in most existing works \cite{struppek2023exploiting, zhang2023iti, luccioni2024stable}.
\end{remark}

\subsubsection{\textbf{Explainability}}
Explainable Artificial Intelligence (XAI) aims to create clear, understandable explanations for AI decisions. In general, XAI methods can be classified from three perspectives \cite{das2020opportunities, arrieta2020explainable}.
\textbf{Scope:} \textit{(1) Local XAI} focuses on explaining individual data instances, such as generating one explanation heatmap \( g \) per instance \( x \in X \). \textit{(2) Global XAI} explains a group of data instances by generating one or more explanation heatmaps. \textbf{Methodology:} \textit{(1) BackPropagation XAI} relies on the analysis of model gradients to interpret decisions. \textit{(2) Perturbation XAI} involves modifying input data and observing the resulting changes in output to understand the decision-making process. \textbf{Usage:} \textit{(1) Intrinsic XAI} refers to AI models that are interpretable by design, such as decision trees or linear regression models, but are not transferable to other architectures. \textit{(2) Post-Hoc XAI} is applied after the model is trained, independent of the model architecture, and interprets decisions without altering the model.

\begin{figure}[h!]
\centering
\includegraphics[width=1\linewidth]{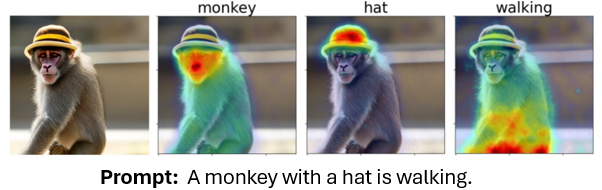}
\caption{Example of Explainability Method: Heatmap from \cite{tang2023daam}.}
\label{fig_explain}
\end{figure}

T2I DMs also suffer from limited interpretability, making it challenging to understand internal mechanisms, which is crucial for further improvements. Studies such as \cite{evirgen2024text, tang2023daam, lee2023diffusion} have undertaken \textit{local} interpretation to explore the explainability of T2I DMs. Fig.~\ref{fig_explain} shows an example from \cite{tang2023daam} explaining T2I DMs by generating an explanation heatmap \( g \) for an instance \( x \in X \), with the definition:
\begin{definition}[Explainability of T2I DMs]
The explainability of T2I DMs focuses on identifying which parts of a generated image are most influenced by specific words. 
\end{definition}

\begin{remark}
\label{remark_def_explainable}
    Traditional XAI offers a variety of methods; however, for T2I DMs, research is still in its early stages. Current work focuses primarily on local XAI for interpreting individual inputs. 
\end{remark}

\subsubsection{\textbf{Privacy}}
Privacy attacks, which aim to reveal information not intended to be shared--such as details about the training data, the model itself, or unintended data biases--are a major concern in traditional DL tasks. Privacy attacks can be broadly classified into two categories: \textit{Training Data Privacy Attacks} and \textit{Model Privacy Attacks} \cite{rigaki2023survey}.

\noindent(1) Training Data Privacy Attacks

\textit{Membership Inference Attacks (MIAs)} aim to determine whether a specific data point \(x\) was included in a model's training set \cite{shokri2017membership, hu2022membership, liu2021machine}. They typically require the adversary to have prior knowledge of the target data.

\textit{Data Extraction Attacks} allow an adversary to directly reconstruct sensitive information from the model using only query access \cite{carlini2021extracting, balle2022reconstructing}.

\textit{Property Inference Attacks} occur when attackers discover hidden properties about training data that are not included as features \cite{ganju2018property, staab2024beyond}. E.g., they might determine the gender ratio in a dataset even if it is not recorded.

\noindent(2) Model Privacy Attacks

\textit{Model Extraction Attacks} involve an adversary attempting to extract information from the target model and recreate it by building a substitute model \(\hat{f}\) that mimics the original model \(f\), replicating its functionality without accessing its architecture or parameters \cite{Krishna2020Thieves, jagielski2020high}.

In addition to privacy attacks, another major privacy concern is the \textit{Memorization Phenomenon} \cite{carlini2019secret, carlini2023quantifying}, which refers to the tendency of models to memorize and reproduce training data, especially when the training data contains sensitive or copyrighted material. 

T2I DMs are increasingly susceptible to various \textit{Privacy Attacks} and \textit{Memorization Phenomenon}. A series of studies have explored MIA, data extraction attacks \cite{carlini2023extracting, li2024shake}, and the memorization phenomenon \cite{wen2024detecting, ren2024unveiling, somepalli2023diffusion, somepalli2023understanding, carlini2023extracting} in T2I DMs, offering tailored definitions of these vulnerabilities in the context of T2I models.




\begin{definition}[Membership Inference Attacks for T2I DM]
Given a T2I DM \(f_\theta\) parameterised by weight \(\theta\) and dataset \(\mathcal{D} = \{(x_1,y_1), (x_2,y_2), \ldots, (x_n,y_n)\}\), where \(\mathcal{D} = \mathcal{D}_M \cup \mathcal{D}_H\) and \(f_\theta\) is trained on \(\mathcal{D}_M\), called the member set, and \(\mathcal{D}_H\) is the hold-out set \cite{carlini2022membership, duan2023diffusion}. One typical white-box MIA is defined using a loss-based thresholding rule:
\begin{equation}\label{MIA_threshold_Attack}
    \mathcal{M}(x_i,\theta) = I [\mathcal{L}(x_i, \theta) < \gamma],
\end{equation}
where \( \mathcal{M}(x_i, \theta) \in \{0,1\} \) is the attacker's prediction: it outputs 1 if \( (x_i,y_i) \in \mathcal{D}_M \), and 0 otherwise. \( \mathcal{L}(x_i, \theta) \) denotes the loss function, and \( \gamma \) is a given threshold.
\end{definition}

\begin{definition}[Data Extraction Attack for T2I DM]
\label{def_data_extraction}
    An image \(y\) is extracted from a DM \(f_\theta\) if there exists an attacking algorithm \(\mathcal{A}\) such that \(\hat{y} = \mathcal{A}(f_\theta)\) has the property that \(d(y, \hat{y}) \leq \delta\), where \(d\) is a distance function (Euclidean \(l_2\)-norm distance) and \(\delta\) is threshold that determined whether two images are identical.   
\end{definition}

\begin{figure}[h!]
\centering
\includegraphics[width=0.7\linewidth, height=0.3\linewidth]{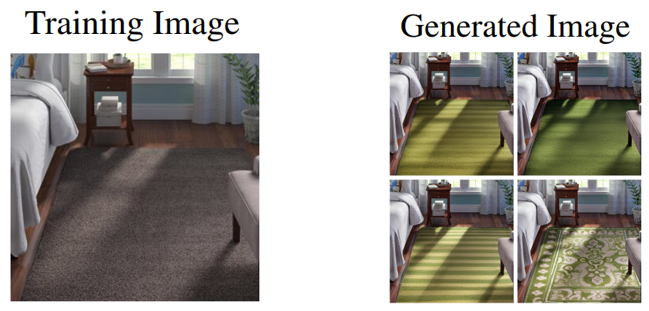}
\caption{Example of \textit{memorization phenomenon} in T2I DMs giving prompt ``Plattville Green Area Rug by Andover Mills'' from \cite{wen2024detecting}.}
\label{fig_memorization}
\end{figure}

A successful MIA can identify training samples within the distribution, enabling an adversary to extract generated outputs that are likely derived from the original training data, thus facilitating data extraction attacks. Furthermore, all privacy attack methods can be adopted to exploit the \textit{memorization phenomenon}, which measures the tendency of T2I DMs to memorize and reproduce training data, as shown in Fig.~\ref{fig_memorization}.

\begin{remark}
\label{remark_def_privacy}
    Research on privacy in T2I DMs has mainly focused on memorization \cite{ren2024unveiling, wen2024detecting, jeon2024understanding, chen2025exploring}, MIAs \cite{zhai2024membership,  li2024unveiling} and data extraction \cite{carlini2023extracting}, revealing risks of reproducing sensitive data \cite{somepalli2023diffusion, somepalli2023understanding}. However, other privacy threats like model extraction and property inference attacks have received less attention. This gap may stem from the complex training objectives and multi-modal nature of T2I DMs \cite{zelaszczyk2024text, 10481956, peng2024comparative, muller2023multimodal}. 
\end{remark}


\subsubsection{\textbf{Factuality}}
While the rapid rise of generative AI models like ChatGPT and SD has revolutionised content creation, hallucination has become a significant trustworthy concern \cite{bang-etal-2023-multitask}. Hallucination refers to the phenomenon where the model generates nonfactual or untruthful information \cite{lee2022factuality, guan2024hallusionbench}, which can be classified into four types based on the modalities \cite{yu2024fake}:

\textbf{(1) Text Modality:}  Generated by LLMs, resulting in fabricated text like fake news \cite{su-etal-2022-read}.

\textbf{(2) Audio Modality:} Created using Deepfake technologies \cite{almutairi2022review}, involving text-to-speech, voice conversion.

\textbf{(3) Visual Modality:} Leveraging DMs to generate or alter images, distorting reality \cite{malik2022deepfake}.

\textbf{(4) Multi-modal:} Arising in systems that combine text, image, and audio inputs, potentially leading to misalignment between modalities \cite{liz2024generation}.

\begin{figure}[h!]
\centering
\includegraphics[width=1\linewidth]{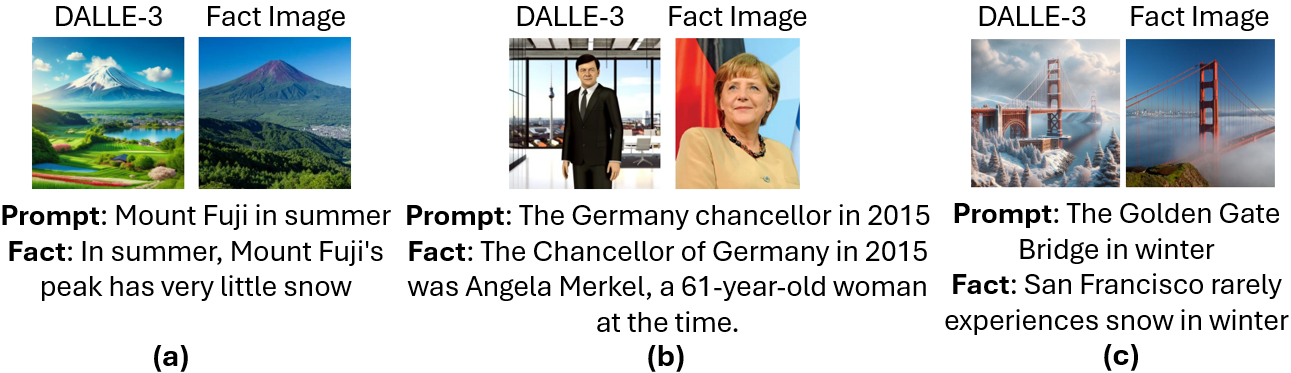}
\caption{Example of (a) \textit{Factual Inconsistency}; (b) \textit{Outdated Knowledge Hallucination}; (c) \textit{Factual Fabrication} in T2I DMs from \cite{lim2024addressing}.}
\label{fig_hallucination}
\end{figure}

In the context of T2I DMs, factuality may specifically refer to the generation of factually inconsistent images, where the output fails to align with the factual information, often termed as \textit{image hallucination}. Based on LLM research \cite{huang2023survey}, Lim et al. \cite{lim2024addressing} define:
\begin{definition}[Hallucination in T2I DMs]
\label{def_Factuality}
In T2I DMs, hallucination occurs when the generated image fails to align with the common sense or facts described by the text, rather than simply being a mismatch with the text prompt. Fig.~\ref{fig_hallucination} shows three representative types of hallucination:
\textit{Factual Inconsistency}: arises from co-occurrence bias; \textit{Outdated Knowledge Hallucination}: the model does not reflect current information; \textit{Factual Fabrication}: the generated image has little to no basis in reality. 

\end{definition}

\subsection{Means}
\label{sec_means}
Section \ref{sec_prop} defines key non-functional properties of trustworthy T2I DMs. This section will offer a detailed review of means designed to study these properties. 

\subsubsection{\textbf{Falsification}}
\label{sec_Falsification}
The straightforward approach to revealing vulnerabilities is falsification, which aims to test hypotheses and identify conditions under which they prove false.

\noindent \textbf{Falsification for Robustness:} 
Most preliminary robustness studies aim to design intricate attacks on the model to demonstrate its flaws or weaknesses. 

Zhuang et al. \cite{zhuang2023pilot} show that the vulnerability of T2I DMs stems from the text encoders by attacking the input text $x$. They propose three attack methods (PGD, greedy search and genetic algorithms) to generate $x'$ by the optimization:
\begin{equation}\label{eq_attack_clip}
\min_{x'} cos(\tau_\theta(x), \tau_\theta(x')),
\end{equation}
where \( \tau_\theta(x) \) denotes the text encoder of CLIP and \(cos\) refers to the cosine similarity. The goal is to find a perturbed example \(x'\) that is semantically different from \(x\), thereby causing the model to generate incorrect content.

Liu et al. \cite{liu2023riatig} propose a similar method, RIATIG, which aims to create an AE \( x' \) that generates a semantically similar image to \( y \), while ensuring that \( x' \) is sufficiently different from \( x \) to avoid detection. They employ a genetic algorithm to find \(x'\), for a T2I DM \(\mathcal{G}\), the optimization is:
\begin{equation}
\arg\max_{x'} S(\mathcal{G}(x'), y), \, \text{ s.t. } d(x, x') > \gamma,
\end{equation}
\noindent where \( S \) is a semantic similarity function of images and \(\gamma\) is the threshold of the semantic dissimilarity of \(x \) and \(x'\).

Du et al. \cite{du2024stable} also design an attack, targeting input text, to identify the prompt $x'$ which leads to a 3rd party vision model \( h : Y \to N \) failing to predict the desired class \( n \), i.e., $\arg\max_i h(\tilde{y})_i \neq n$. Given the loss function \(\mathcal{L}\) of the vision model, the attack objective is:
\begin{equation}
x' = \argmax_{x': d(x,x') \leq \gamma} \mathcal{L}(n, h(\mathcal{G}(x'))),
\end{equation}

Yang et al. \cite{yang2024on} propose MMP-Attack (multi-modal priors-attack), which adds a target object to the image while removing the original object. They design a gradient-based algorithm to minimize the distance between the original prompt and the target category (to add). Given an original prompt \(x\), a target category \(t\) which is irrelevant to \(x\), the cheating suffix \(a\) needs to be optimized to guide the model generate an image containing \(t\) but unrelated to \(x\), the optimization objective is:
\begin{equation}
\arg\max_{a} \mathbb{E}_{y \sim \mathcal{G}(\tau(x \oplus a))} A(y, t, x),   
\end{equation}
where \(y\) represents a randomly generated image based on the prompt \(x \oplus a\) and \(A(\cdot)\) denotes evaluation metrics such as CLIP scores for image-text matching, along with two object detection metrics: Original Category Non-Detection Rate (if the original category is missed) and Target Category Detection Rate (if the target category is present).

Liu et al. \cite{liu2024discovering} propose SAGE, which implements a gradient-guided search (PGD) over the text encoder space (finding adversarial token embedding \(e_{a}\)) and the high-dimensional latent space (finding latent space perturbation \(d_z\)) to discover failure cases in T2I DMs:
\begin{equation}
\begin{aligned}
e_{a} = e_{a} + r \cdot \alpha \cdot \text{sign}(\nabla_{e_{a}} \mathcal{L}(z, \tau_\theta(x \oplus a))), \\
d_{z} = d_{z}  + r \cdot \alpha \cdot \text{sign}(\nabla_{d_{z} } \mathcal{L}(z + d_{z}, \tau_\theta(x))), \\
\end{aligned}
\end{equation}
where $\mathcal{L}(z, \tau_\theta(x \oplus a ))\!\! =\!\! -f(\mathcal{G}(z, \tau_\theta(x \oplus a )))$ and \(\mathcal{L}(z + d_z, \tau_\theta(x)) = -f(\mathcal{G}(z + d_z, \tau_\theta(x)))\). Here, \(f\) is a discriminative classifier (e.g., ViT) used to determine whether the key object is present in the generated image, \(\alpha\) is the step size, and \(r \in [0,1]\) is a random value. For optimizing the token embedding \(e_a\), given the input text \(x\) as ``A photo of a [class]'', the adversarial token a is appended to \(x\), forming \(x \oplus a\). Minimizing the loss involves finding the adversarial token embedding \(e_a\) that causes the model to generate an image with the wrong object, while \(z\) is the fixed latent space during optimization. Similarly, for the latent space perturbation optimization, the goal is to find a small perturbation \(d_z\) for a random latent code \(z\) that leads to a failure in the generation process.

Shahariar et al.~\cite{shahariar-etal-2024-adversarial} study how adversarial attacks targeting different part-of-speech (POS) tags in text prompts affect image generation in T2I diffusion models. Using a gradient-based token search method, they find that nouns, proper nouns, and adjectives are the most vulnerable, while verbs, adverbs, and numerals are largely resistant.

\begin{table}[htbp]
\caption{Overview of robustness in T2I DMs.}
\resizebox{\linewidth}{!}{
\begin{tabular}{c|c|c|c|c|c|c}
\hline
\textbf{Paper} & \textbf{Robustness} & \textbf{Attack objective} & \textbf{Target} & \textbf{Defence} & \textbf{Model} & \textbf{Year} \\ \hline
Gao \cite{gao2023evaluating} & worst-case & text encoder & untargeted & No & \parbox[c]{2cm}{\centering SD \\ DALL-E 2} & 2023 \\ \hline
Zhuang \cite{zhuang2023pilot} & worst-case & text encoder & untargeted \& targeted & No & \parbox[c]{2cm}{\centering SD} & 2023 \\ \hline
Liu \cite{liu2023riatig} & worst-case & text encoder & untargeted & Yes & \parbox[c]{2cm}{\centering Imagen \\ DALL-E} & 2023 \\ \hline
Du \cite{du2024stable} & worst-case & text encoder & untargeted & No & \parbox[c]{2cm}{\centering SD} & 2023 \\ \hline
Zhang \cite{zhang2024protip} & probabilistic & text encoder & untargeted & Yes & \parbox[c]{2cm}{\centering SD} & 2024 \\ \hline
Yang \cite{yang2024on} & worst-case & text encoder & targeted & No & \parbox[c]{2cm}{\centering SD;DALL\(\cdot\)3} & 2024 \\ \hline
Liu \cite{liu2024discovering} & worst-case & text encoder; U-Net & untargeted & No & \parbox[c]{2cm}{\centering SD; GLIDE \\ DeepFloyd} & 2024 \\ \hline

Shahariar \cite{shahariar-etal-2024-adversarial} & worst-case & text encoder & targeted & No & \parbox[c]{2cm}{\centering SD} & 2024 \\ \hline

\end{tabular}
}
\label{table_summary_robustness}
\end{table}

\begin{remark}[Falsification Focus on Text Encoder]
\label{remark_means_robust_falsification}
    Research on the robustness of T2I DMs via falsification has focused exclusively on the text encoder, as shown in Table~\ref{table_summary_robustness}. There is a notable lack of studies examining the vulnerabilities of the diffusion process. This gap may stem from the nature of the diffusion process, which involves typically hundreds of denoising steps, results in the vanishing gradient problem \cite{liu2024discovering}.
\end{remark}

\noindent \textbf{Falsification for Security:} Similar to robustness, many security studies also aim to design sophisticated backdoor attacks on T2I DMs to expose their vulnerabilities.

Struppek et al. \cite{struppek2023rickrolling} propose a teacher-student framework to inject backdoors (non-Latin homoglyph characters) into the text encoder. A poisoned student text encoder $\widetilde{\tau}_\theta$ computes the same embedding for inputs $x$ containing the trigger character $t$, as the clean teacher encoder $\tau_\theta$ does for the prompt $x_t$ that represents the desired target behavior. Additionally, a utility loss is defined to ensure that the $\widetilde{\tau}_\theta$ produces embeddings similar to those of the $\tau_\theta$:
\begin{equation}
\begin{aligned}
\mathcal{L}_\mathit{Backdoor}& = \frac{1}{|X|} \sum_{x \in X} d\left( \tau_\theta(x_t), \widetilde{\tau}_\theta(x \oplus t) \right), \\
\mathcal{L}_\mathit{Utility} &= \frac{1}{|X'|} \sum_{x \in X'} d\left( \tau_\theta(x), \widetilde{\tau}_\theta(x) \right), \\
\mathcal{L} &= \mathcal{L}_\mathit{Utility} + \beta \cdot \mathcal{L}_\mathit{Backdoor},
\end{aligned}
\end{equation}
\noindent where \(d\) indicates a similarity metric, \({X'}\) is different batch from \({X}\) during each training step and the final loss function \(\mathcal{L}\) is weighted by \(\beta\).

Zhai et al. \cite{zhai2023text} propose a multi-modal backdoor attack called BadT2I to target the DM. They design attacks at three levels of vision semantics: Pixel-Level, Object-Level, and Style-Level. E.g., the Pixel-Level backdoor attack aims to tamper with specific pixels in the generated image. They proposed attack loss along with a regularization loss (prevent overfitting to target patches) as:
\begin{equation}
\begin{aligned}
\mathcal{L}_{\mathit{Bkd}\text{-}\mathit{Pix}} &= \mathbb{E}_{z_p, c_{tr}, \epsilon, t} \left[ \left\| \epsilon_{\theta} (z_{p,t}, t, c_{tr}) - \epsilon \right\|_2^2 \right], \\
\mathcal{L}_{\mathit{Reg}} &= \mathbb{E}_{z,c,\epsilon,t} \left[ \left\| \epsilon_{\theta}(z_{t}, t, c) - \hat{\epsilon}(z_{t}, t, c) \right\|^2_2 \right], \\
\mathcal{L}& = \lambda \cdot \mathcal{L}_{\mathit{\mathit{Bkd}\text{-}\mathit{Pix}}} + (1 - \lambda) \cdot \mathcal{L}_{\mathit{Reg}},
\end{aligned}
\end{equation}
where \( z_{p,t} \) is the noisy version of \( z_{p} := \varepsilon(y_{\text{patch}}) \), and \( c_{tr} := \tau_\theta(x_{tr}) \). Here, \( \varepsilon \) is the image encoder, and \( y_{\text{patch}} \) refers to an image with the target patch added, while \( x_{tr} \) denotes the text input containing the trigger \([T]\). \( \hat{\epsilon} \) represents a frozen pre-trained U-Net. The overall loss function \(\mathcal{L}\) is weighted by \( \lambda \in [0, 1] \). The model then generates images containing a pre-set patch whenever the inputs include the trigger. Object- and style-level backdoor attacks follow the similar loss objective but target specific differences.

Vice et al. \cite{vice2024bagm} later exploit invisible triggers instead of previous visible triggers like non-Latin characters, by embedding the trigger as part of the original prompt \(x\) (cf. Sec.\ref{Sec_security_def}). They propose BAGM  to attack three stages of T2I DMs: tokenizer, text encoder and diffusion components (U-Net). This approach includes surface attacks (involving appending, replacing, and prepending methods applied to the tokenizer stage), shallow attacks (fine-tuning the text encoder with poisoned data), and deep attacks (fine-tuning the U-Net while keeping all text-encoder layers frozen). 

Huang et al. \cite{huang2024personalization} investigate the implanting of backdoors through personalization methods (Textual Inversion and DreamBooth). They demonstrated that the backdoor can be established by using only 3-5 samples to fine-tune the model and explored implanting visible triggers during the fine-tuning phase.


Traditional backdoor attacks require extensive data and training to fine-tune victim models. Wang et al. \cite{wang2024eviledit} introduce EvilEdit, a training- and data-free model editing-based backdoor attack. They directly modify the projection matrices in the cross-attention layers to align the projection of the textual trigger with the backdoor target. Given a trigger $x_{tr}$ and a backdoor target $x_{ta}$, the goal is to manipulate the model so that the image generated by $x \oplus x_{tr}$ matches the description of $x \oplus x_{ta}$, where $x$ is the original prompt:
\begin{equation}
\begin{aligned}
    \mathcal{G}^* = \arg \min_{\mathcal{G}^*} \| \mathcal{G}^*(x \oplus x_{tr}) - \mathcal{G}(x \oplus x_{ta}) \|_2^2, \\
    \| \mathbf{W}^* \mathbf{c}_{tr} - \mathbf{W} \mathbf{c}_{ta} \|_2^2 < \gamma,
\end{aligned}
\end{equation}
where $\mathcal{G}$ and $\mathcal{G}^*$ denote the clean and backdoored T2I DMs, respectively. The alignment of the projections is achieved by modifying the projection matrices \(\mathbf{W}\) and \(\mathbf{W}^*\), representing the clean and backdoored projection matrices, respectively. The projections of the trigger embeddings $\mathbf{c}_{tr} \! = \!\tau_\theta(x_{tr})$ and the backdoor target $\mathbf{c}_{ta}\! =\! \tau_\theta(x_{ta})$ are considered aligned if their distance is less than a threshold~$\gamma$.

Naseh et al.~\cite{naseh2024injecting} designed a backdoor attack to inject specific biases into T2I models while minimizing their impact on model utility. The framework consists of three main stages:
1) \textbf{Trigger-Bias Selection:} Select two trigger tokens (a noun and a verb/adjective) as bias triggers. 2) \textbf{Poisoned Sample Generation:} Construct a dataset that includes both clean and  subtly triggered biased samples. 3) \textbf{Bias Injection (Fine-tuning):} Fine-tuning the model on this dataset to activate biased outputs only when triggers are present.

\begin{remark}
\label{remark_means_security_falsification}
    Previous backdoor studies have targeted the main components of T2I DMs: the tokenizer, text encoder, and denoising model, as shown in Table~\ref{table_security}. However, several limitations remain: \textbf{(1)} Current triggers are inflexible, with fixed patterns and locations \cite{vice2024bagm, naseh2024injecting}. Future research should explore dynamic triggers \cite{zheng2022detecting, salem2022dynamic}, which can show random patterns and locations. \textbf{(2)} Existing works focus on poisoning-based \cite{struppek2023rickrolling, zhai2023text, huang2024personalization, vice2024bagm} and weights-oriented attacks \cite{wang2024eviledit}, but no study has explored structure-modified backdoors, where hidden backdoors are added by changing the model's structure, as seen in traditional DL systems.
\end{remark}


\begin{table}[htbp]
\caption{Overview of backdoor attacks in T2I DMs.}
\resizebox{\linewidth}{!}{
\begin{tabular}{c|c|c|c|c}
\hline
\textbf{Paper} & \textbf{Attack Objective} & \textbf{Defence} & \textbf{Model} & \textbf{Time} \\ \hline
Struppek \cite{struppek2023rickrolling}  & text-encoder & No & \parbox[c]{3cm}{\centering SD} & 2023 \\ \hline
Zhai \cite{zhai2023text}  & U-Net & No & \parbox[c]{3cm}{\centering SD} & 2023 \\ \hline
Huang \cite{huang2024personalization}  & text-encoder; U-Net & No & \parbox[c]{3cm}{\centering DreamBooth; Textual Inversion} & 2024 \\ \hline
Wang \cite{Wang2024T2IShield}  & text-encoder; U-Net & Yes & \parbox[c]{3cm}{\centering SD} & 2024 \\ \hline
Vice \cite{vice2024bagm}  & tokenizer; text-encoder; U-Net & No & \parbox[c]{3cm}{\centering SD; Kandinsky; DeepFloyd-IF} & 2024 \\ \hline

Wang \cite{wang2024eviledit}  & text-encoder; U-Net & No & \parbox[c]{3cm}{\centering SD} & 2024 \\  \hline

Chew \cite{chew2024defending} & text-encoder & Yes & \parbox[c]{3cm}{\centering SD} & 2024 \\ 
\hline

Naseh \cite{naseh2024injecting} & text-encoder & No & \parbox[c]{3cm}{\centering SD} & 2025 \\ \hline

Guan \cite{guan2025ufid} & text-encoder; U-Net & Yes & \parbox[c]{3cm}{\centering SD} & 2025 \\ \hline

\end{tabular}
}
\label{table_security}
\end{table}

\noindent \textbf{Falsification for Privacy:}
Privacy attacks, such as MIA and data extraction attacks, are applied to T2I DMs to expose their privacy vulnerabilities. Additionally, various works aim to design detection algorithms to investigate the memorization phenomenon in T2I DMs. Table~\ref{table_privacy} outlines the existing privacy studies, as discussed in Remark~\ref{remark_def_privacy}.

Duan et al. \cite{duan2023diffusion} proposed Step-wise Error Comparing Membership Inference (SecMI), a query-based MIA relying on the error comparison of the forward process posterior estimation based on the common overfitting assumption in MIA where member samples (\(x_m \in \mathcal{D}_M\)) have smaller posterior estimation errors, compared with hold-out samples (\(x_h \in \mathcal{D}_H\)). The local estimate error of single data point \(x_0\) at timestep \(t\) is:
\begin{equation}
\begin{aligned}
    \ell_{t,x_0} &= \| \hat{x}_{t-1} - x_{t-1} \|^2, \\
    \ell_{t, x_m} &\leq \ell_{t, x_h}, \quad 1 \leq t \leq T,
\end{aligned}
\end{equation}
\noindent where $x_{t-1} \sim q(x_{t-1} \mid x_t, x_0)$, $\hat{x}_{t-1} \sim p_\theta(\hat{x}_{t-1} \mid x_t)$.

Dubinski et al. \cite{dubinski2024towards} execute MIA on a new dataset, LAION-mi, in three scenarios: black-box (access to input and output), grey-box (access to visual and text encoders), and white-box (access to trained weights). They then apply threshold-based MIA cf.~Def.~\ref{MIA_threshold_Attack} using two error metrics: \textbf{Pixel Error}, the distance between the original image \(y\) and the generated image \(y'\); and \textbf{Latent Error}, the difference between the latent representations of \(y\) and \(y'\). An image \(y\) is classified as a member if the combined error \(\mathcal{L}\)  falls below a predefined threshold \(\gamma\).

Li et al.~\cite{li2024unveiling} propose a structure-based MIA for T2I DMs by comparing structural differences between members and non-members. Specifically, the latent representation \(z_0\) of an image \(x_0\) is corrupted using the DDIM process to produce \(z_t\). The decoder of the T2I DM is then used to reconstruct \(z_t\) back into the pixel space, yielding a reconstructed image \(x_t\). The Structural Similarity Index (SSIM) between \(x_0\) and \(x_t\) is computed to define a membership score, based on which the membership status is predicted as follows:
\begin{equation}
\label{eq:ssim_mia}
x_0 =
\begin{cases}
\text{member}, & \text{if } \text{SSIM}(x_0, x_t) > \gamma \\
\text{non-member}, & \text{if } \text{SSIM}(x_0, x_t) \leq \gamma
\end{cases}
\end{equation}

Zhai et. al.~\cite{zhai2024membership} proposed Conditional Likelihood Discrepancy (CLiD) to perform MIA. They first identified the assumption that T2I DMs overfit more to the conditional distribution \(\log p(x\!\! \mid\!\! c)\) than to the marginal distribution \(\log p(x)\). Based on this, they derive a sample-wise membership inference indicator and employ KL divergence as the distance metric, formulated as:
\begin{align}
\mathbb{E}_{q_{\text{mem}}(x, c)} \left[ \log p(x \mid c) - \log p(x) \right] 
& \geq \\ \mathbb{E}_{q_{\text{out}}(x, c)} \left[ \log p(x \mid c) - \log p(x) \right] \nonumber
& + \delta_H.
\end{align}
Then given \( I(x, c)\!\! = \!\!\log p(x \!\!\!\!\mid\!\!\!\! c)\! - \!\log p(x) \), it holds that \( \mathbb{E}_{q_{\text{mem}}}[I(x)] \geq \gamma \geq \mathbb{E}_{q_{\text{out}}}[I(x)] \). As an approximation of the log-likelihoods, the indicator \( I(x, c) \) is defined as:
\[
I(x, c) = \mathbb{E}_{t, \epsilon} \left[ \left\| \epsilon_\theta(z_t, t \mid c_\emptyset) - \epsilon \right\|^2 \right] 
- \mathbb{E}_{t, \epsilon} \left[ \left\| \epsilon_\theta(z_t, t\mid c) - \epsilon \right\|^2 \right],
\]
where \( c_\emptyset \) denotes an empty text condition input used to estimate the approximation of \( \log p_\theta(x) \).  If \( I(x, c) \) exceeds a threshold \( \gamma \), the instance is likely from \( q_{\text{mem}} \).

Carlini et al. \cite{carlini2023extracting} introduce a black-box data extraction attack on T2I DMs. Their approach involves generating numerous images and then applying a distance-based MIA to estimate generation density, based on the Def.~\ref{def_data_extraction}, as adopted from \cite{carlini2021extracting}. They observe that memorized prompts tend to generate nearly identical images across different seeds, resulting in high density.

Li et al. \cite{li2024shake} reveal that fine-tuning pre-trained models with manipulated data can amplify privacy risk on DMs. They design  Shake-To-Leak, a pipeline that fine-tunes T2I DMs in three steps: 1) generate a synthetic fine-tuning dataset from a T2I DM using a target prompt; 2) conduct fine-tuning; 3) perform MIA and data extraction. Their empirical results demonstrate that fine-tuning on manipulated data increases the likelihood of leaking information from the original pre-training set, showing that the T2I synthesis mechanism can be exploited to extract sensitive content.

Another stream of work focused on the memorization phenomenon. Wen et al. \cite{wen2024detecting} study memorized prompts in T2I DMs by introducing a detection method based on
the magnitude of text-conditional noise predictions, leveraging the observation that memorized prompts exhibit stronger
text guidance. Therefore, given a prompt embedding \(e_p\) and sampling step \(T\), they define the  text-conditional noise prediction as the detection metric:
\begin{equation}
d = \frac{1}{T} \sum_{t=1}^{T} \|\epsilon_\theta(z_t,t \mid e_p) - \epsilon_\theta(z_t,t \mid {e_\emptyset})\|^2,
\end{equation}
where \(e_\emptyset\) is the prompt embedding of an empty string.


Similarly, Ren et al. \cite{ren2024unveiling} investigate the memorization in T2I DMs by analyzing cross-attention mechanisms. They observe that memorized prompts exhibit anomalously high attention scores on specific tokens and propose attention entropy to quantify the dispersion of attention:
\begin{equation} \label{entropy_attention}
    \quad E_t = \sum_{i=1}^N -\overline{a}_i \log(\overline{a}_i), 
\end{equation}
where \( N \) is the number of tokens, \( t \) is the diffusion step, and \( \overline{a}_i \) is the average attention score for the \( i \)-th token. Higher entropy indicates more concentrated attention and serves as a signal of potential memorization.

Somepalli et al. \cite{somepalli2023diffusion} identify memorized prompts by directly comparing the generated images with the original training data. They defined replication informally, stating that a generated image is considered to have replicated content if it contains an object that appears identically in a training image. They also found that text conditioning is a major factor in data replication \cite{somepalli2023understanding}.

Jeon et al. \cite{jeon2024understanding} observe that memorization correlates with
regions of sharpness in the probability landscape, which can
be quantified via the Hessian of the log probability, denote as \( H_{\theta}(\bm{x}_t)\! :=\!\! \nabla^2_{\bm{x}_t} \log p_t(\bm{x}_t) \) for the unconditional case, and \( H_{\theta}(\bm{x}_t, c)\! := \!\!\nabla^2_{\bm{x}_t} \log p_t(\bm{x}_t\!\! \mid \!\! c) \) for the conditional case, where \(p_t(\bm{x}_t)\) denotes the marginal data distribution at timestep \(t\). Large negative eigenvalues of the Hessian indicate sharp, isolated regions in the learned distribution, suggesting memorization of specific data points and providing a mathematically grounded explanation for memorization. 

Chen et al. \cite{chen2025exploring} discuss the difference between global memorization (the entire training image is memorized) and local memorization (only parts of the training image are memorized), and show that current detection metrics can be easily tricked in local memorization scenarios due to diversity in non-memorized regions, resulting in many false negatives during evaluation. Thus, they refine the metric proposed by Wen et al. \cite{wen2024detecting} for
local memorization by incorporating end-token masks that empirically highlight locally memorized regions as:
\begin{equation}
\label{eq:ld}
LD \!\!=\!\! \frac{1}{T} \sum_{t=1}^{T} \left\| \left( \epsilon_\theta(z_t,t \!\!\mid \!e_p)\! -\! \epsilon_\theta(z_t,t \!\mid \!\!{e_\emptyset}) \right) \circ \bm{m} \right\|_2\!\!\! 
\left/\!\! 
\left( \frac{1}{N} \sum_{i=1}^{N} m_i \right)
\right.
\end{equation}
Here, \(N\) is the number of elements in the mask \(\bm{m}\), and \(\circ\) denotes element-wise multiplication.

\begin{table}[htbp]
\caption{Overview of privacy in T2I DMs.}
\resizebox{\linewidth}{!}{
\begin{tabular}{c|c|c|c|c}
\hline
\textbf{Paper} & \textbf{Privacy} & \textbf{Defense} & \textbf{Model} & \textbf{Time} \\ \hline
Duan \cite{duan2023diffusion}  & MIA & Yes & \parbox[c]{2cm}{\centering SD; LDM} & 2023 \\ \hline
Carlini \cite{carlini2023extracting}  & Data Extraction Attack & Yes & SD; Imagen & 2023 \\ \hline
Somepalli \cite{somepalli2023diffusion}  & Memorization & No & SD & 2023 \\ \hline
Somepalli \cite{somepalli2023understanding}  & Memorization & Yes & SD & 2023 \\ \hline
Ren \cite{ren2024unveiling}  & Memorization & Yes & SD & 2024 \\ \hline
Dubinski \cite{dubinski2024towards}  & MIA & No & SD & 2024 \\ \hline
Wen \cite{wen2024detecting}  & Memorization & Yes & SD & 2024 \\ \hline
Li \cite{li2024shake}  & MIA; Data Extraction Attack & No & SD & 2024 \\ \hline 

Li \cite{li2024unveiling}  & MIA & No & SD & 2024 \\ \hline 

Zhai ~\cite{zhai2024membership} & MIA & No & SD & 2024 \\ \hline 

Jeon \cite{jeon2024understanding}  & Memorization & Yes & SD & 2025 \\ 
\hline
Chen \cite{chen2025exploring}  & Memorization & Yes & SD & 2025 \\ 
\hline
Chen \cite{Chen_2025_CVPR}  & Memorization & Yes & SD & 2025 \\ 
\hline
\end{tabular}
}
\label{table_privacy}
\end{table}

\noindent \textbf{Falsification for Fairness:}  
Some work also try to injects bias into DMs to compromise fairness. Huang et al.~\cite{huang2025implicit} propose IBI-Attacks, an implicit bias injection framework that computes a general bias direction in the prompt embedding space and adaptively modifies it based on the input. The method consists of three stages: (1) \textbf{Bias direction calculation:} A LLM is used to rewrite prompts according to specific bias. The bias direction vector is then calculated as the average difference between the embeddings of the original and rewritten prompts. 
(2) \textbf{Adaptive feature selection:} An input-dependent module is trained to adjust the fixed bias direction based on the user's prompt. (3) \textbf{User inference:} The adapted bias direction is injected into the user’s prompt embedding to influence the model output accordingly.

\subsubsection{\textbf{Verification \& Validation}}
Falsification is understood as the refutation of statements, whereas verification refers to statements that are shown to be true \cite{fretwurst2017verification}. V\&V entails confirming the correctness or effectiveness of a hypothesis or model. 

\noindent \textbf{V\&V for Robustness:}
Zhang et al. \cite{zhang2024protip} first propose a verification framework, ProTIP, to evaluate the probabilistic robustness of T2I DMs as defined in Def. \ref{def_Prob_rob_t2i}. They applied this framework to verify the probabilistic robustness of existing open-source T2I DMs SD.

\begin{remark}\label{remark_means_robust_VV}
    Among all the properties and their addressing means, V\&V work is relatively underexplored. Only one study \cite{zhang2024protip} has proposed a framework to verify the robustness of T2I DMs. This is largely because defining a specification for verification is challenging. For example, fairness studies are often done case-by-case due to the variety of biases (e.g., gender bias is binary, while racial bias is multi-class), making it difficult to design a general verification specification. Similarly, in security, backdoor attacks are designed to be subtle, with triggers constantly changing, making it difficult to establish a unified framework for verifying all potential security attacks.
\end{remark}

\subsubsection{\textbf{Assessment}}
Assessment typically involves designing intricate  metrics to assess specific attributes of a model without targeting a specific predefined specification.

\noindent \textbf{Assessment for Fairness:} Assessment is often employed in fairness studies to assess the extent of bias in a T2I DM. Based on the assessment results, corresponding mitigation efforts can be applied.

Struppek et al. \cite{struppek2023exploiting} propose VQA Score and Relative Bias to measure the cultural biases induced by homoglyhps. The VQA Score is used to measure how much homoglyphs introduce cultural bias. They feed the generated images into BLIP-2 \cite{li2023blip} and ask if the model detects specific cultural traits. For example, to check if an African homoglyph affects the appearance of people, they ask: ``Do the people shown have an African appearance?''. Then the VQA Score is the ratio in which the model answers `yes'. The Relative Bias measures how much a single non-Latin character shifts the image generation towards its associated culture. It quantifies the relative increase in similarity between the given prompt \(x_i\) (explicitly states the culture) and the generated images \(y_i\) and \(\tilde{y_i}\) (with and without the non-Latin character) included in the text prompt), averaged over \(N\) prompts.
\begin{equation}
\begin{aligned}
\text{VQA Score} = \frac{1}{N} \sum_{i=1}^{N} \mathcal{I}[C(y_i, q) = \text{yes}], \\
\text{Relative Bias} = \frac{1}{N} \sum_{i=1}^{N} \frac{S_c(\tilde{y_i}, x_i) - S_c(y_i, x_i)}{S_c(y_i, x_i)},
\end{aligned}
\end{equation}
where \( C(y, q) \) is the answer from the BLIP-2 for image \( y \) and question \( q \). The indicator function \( \mathcal{I} \) returns 1 if the answer is ``yes''. \(S_c\) is the cosine similarity between CLIP embeddings of image \(y\) and text prompt \(x\) .

Friedrich et al. \cite{friedrich2023fair} introduce FairDiffusion to detect biases in SD. They analysis dataset bias by examining the co-occurrence of a biased attribute (e.g., gender) with a target attribute (e.g., occupation). If the proportion of genders for a particular occupation deviates from the fairness definition in Def.~\ref{def_fairness}, it indicates a bias source.

Bansal et al. \cite{bansal2022well} introduce ENTIGEN, a benchmark dataset design to evaluate how image generation changes with ethical text interventions related to gender, skin color, and culture. They assess the diversity of generated images by using diversity scores, CLIP scores, and human evaluations when inputting ethically biased prompts. For example, the diversity score for axis \( g \) (gender) across its groups for category \( P \) is given by:
\begin{equation}
    \mathcal{}{diversity}_{P}^g = \frac{\sum_{k \in P} |s_{k,a}^g - s_{k,b}^g|}{\sum_{k \in P} (s_{k,a}^g + s_{k,b}^g)},
\end{equation}
\noindent where \( s_{k,a}^g \) and \( s_{k,b}^g \) represent the number of images associated with the two groups \( a \) (man) and \( b \) (woman), respectively, across a specific social axis \( g \) (gender).

Luccioni et al. \cite{luccioni2024stable} use captions and open-ended Visual Question Answering (VQA) models to generate textual descriptions of images. They measure the likelihood of gender-marked words (e.g., `man', `woman') or gender-unspecified descriptors (e.g., `person', the profession name) appearing in these descriptions. Their work contribute a dataset of identity and social attributes and a low-code interactive platform for exploring biases.

Zhou et al.\cite{zhou2024association} focus on association-engendered stereotypes—biases arising from the combination of multiple concepts (e.g., while “Black people” and “houses” separately do not imply bias, their association can lead to stereotypical depictions: “Black people in poorer houses”). They propose Stereotype-Distribution-Total-Variation (SDTV), a metric that evaluates stereotype in T2I DMs by measuring the total variation distance between their stereotype probability distributions. They define a probability distribution function \( p(s\!= \!v(s)\! \mid \!y) \) to quantify the likelihood that an object in the image \(y\) exhibits a specific sensitive attribute \(s\) with value \(v(s)\). Given a prompt \(x\), the probability that the generated image \(y\) exhibits sensitive attribute \(s_y\) taking value \(v(s_y)^i\) is denoted as \( p_\theta(s_y \!=\! v(s_y)^i \mid y, x) \). The SDTV score for model \(\mathcal{G}\) is then calculated as:
\begin{equation}
\text{SDTV}(\mathcal{G})\!\! =\!\! \max_{\{i, j\} \subseteq \mathcal{S}}\! \left| p_\theta(s_y\!\! =\!\! v(s_y)^i\!\! \mid \!\! y, x) - p_\theta(s_y\!\!=\!\! v(s_y)^j \!\mid \!y, x) \right|
\end{equation}
where \(\mathcal{S}\) is the set of all possible values for the sensitive attribute. A higher SDTV indicates a stronger stereotype,  as the sensitive attributes represented in the generated images are distributed less evenly.

Chinchure et al.~\cite{chinchure2024tibet} propose TIBET, a T2I Bias Evaluation Tool design to detect and quantify biases of T2I DMs. TIBET follows a three-step pipeline for a given prompt \( x \): (1) A LLM (GPT-3) is used to dynamically identify bias-axes and generate counterfactual prompts for each axis; (2) Image sets for initial prompts (\(I^{i}\)) and the counterfactual prompts (\(I^{cf}\)) are generated using a black-box SD model; (3) \(I^{i}\) and \(I^{cf}\) are then compared to determine whether the images generated from the original prompt exhibit bias toward a specific counterfactual. This is done using VQA-based concept extraction and vision-language embedding models. Specifically, for vision-language embedding models method, all images are embedded using CLIP, and the cosine similarity is computed between \(I^{i}\) and \(I^{cf}\). The result score, \(CAS^{CLIP}\), is defined as the mean cosine similarity between all image pairs:
\begin{equation}
\text{CAS}^{\text{CLIP}} = \text{mean} \left( 
\cos\left(\text{CLIP}(I^i), \text{CLIP}(I^{cf})\right) 
\right),
\end{equation}
CAS values range from 0 to 1, with higher values indicating greater similarity between the two concept sets, thereby suggesting the presence of bias.

Similarly, D’Inca et al.~\cite{d2024openbias} use a LLM combined with a VQA model for open-set bias detection in T2I DMs. Their approach involves three main steps:
\begin{align}
& \text{Build knowledge base:} \quad \mathcal{B}  = \text{LLM}(\mathcal{X}), \\
& \text{Generate images:} \quad \mathcal{I}^x_b = \{ \mathcal{G}(x, s) \mid s \in S \}, \\
& \text{Predict classes:} \quad \hat{c} = \mathtt{VQA}(I, q, \mathcal{C}_b),
\end{align}
where \(\mathcal{X}\) is the dataset of captions, \(\mathcal{B} \) is the knowledge base of biases generated by the LLM from \(\mathcal{X}\), \(b \in \mathcal{B} \) is a bias type, \(x \in \mathcal{X}_b\) is a caption associated with bias \(b\), \(S\) is a set of sampled noise, \(I \in \mathcal{I}^x_b\) is a generated image, \(q\) is a question from the pair \((x, q) \in \mathcal{D}_b\), and \(\mathcal{C}_b\) is the set of possible classes related to the bias.
First, the LLM generates bias for each caption in \(\mathcal{X}\), including bias names, related classes, and bias-detection questions, forming the knowledge base \(\mathcal{B}\). Next, for each bias \(b\) and caption \(x\), a set of images \(\mathcal{I}^x_b\) is generated by varying noise \(s\). Finally, a VQA model predicts bias-related classes \(\hat{c}\) for each image-question pair to analyze the distribution of classes and quantify the bias severity.


\begin{remark}
\label{remark_means_fairness_assess}
    We found that assessment methods have primarily been applied to the study of fairness because fairness involves multiple types of bias from different aspects, such as dataset and model, making the assessment complex. In contrast, for other properties like security, assessment is simpler and often uses straightforward metrics like ASR (Attack Success Rate), which measures the proportion of successful attacks.
\end{remark}

\subsubsection{\textbf{Enhancement}}
Enhancement involves implementing measures to protect a model from adversarial attacks (robustness), data poisoning (security), data de-duplication (privacy) or other threats that impact trustworthiness and performance. 

\noindent \textbf{Enhancement for Robustness:} Some preliminary works have used existing spellchecker tools to defend against these textual perturbations. 

Zhang et al. \cite{zhang2024protip} employ three spell-checking tools to defend against stochastic perturbations in text inputs and evaluate them using their proposed V\&V framework, ProTIP. Similarly, Liu et al. \cite{liu2023riatig} utilize \textit{Grammarly} to mitigate those stochastic perturbation in text input.

\begin{remark}
\label{remark_means_robust_enhance}
    Most research on robustness primarily focuses on falsification to demonstrate the vulnerability of T2I DM. However, there is a limited focus on defense measures. Furthermore, existing defense strategies predominantly rely on spellchecker tools, and there is a lack of efforts aimed at enhancing the internal robustness of the model itself, e.g., by adversarial training.
\end{remark}

\noindent \textbf{Enhancement for Security:} 
Wang et al. \cite{Wang2024T2IShield} propose T2IShield, the first defense method against backdoor attacks. They detect backdoor samples by analyzing cross-attention in the U-Net (cf.~Def.~\ref{U-Net_cross_attention}), finding that the backdoor trigger suppresses other token representations to generate specific content. Therefore, the attention map between a prompt containing the backdoor trigger and one without it will show significant differences. For tokens of length \(L\), the model produces a group of cross-attention maps of the same length \( M \!\!=\!\! \{M^{(1)}, M^{(2)}, \ldots, M^{(L)}\} \), where \(M^{(i)}\!\! =\!\! \frac{1}{T}\sum_{t=1}^T M_t^{(i)}\) is the average cross-attention map over time steps \( T \) and \( i \in [1, L] \). They introduce F-Norm Threshold Truncation (FTT), a statistical method that uses the Frobenius norm \cite{horn2012matrix} to assess the magnitude of a matrix, and Covariance Discriminative Analysis (CDA), another statistical method that leverages the covariance matrix to differentiate between normal and anomalous patterns for detecting backdoor samples.

Chew et al.~\cite{chew2024defending} demonstrate that simple textual perturbations can defend against SOTA backdoor attacks in T2I models, with minimal impact on image quality. They apply semantic-preserving transformations at both the word level (synonym replacement and translation) and character level (homoglyph substitution and random perturbation). These perturbations are applied probabilistically to preprocess prompts, aiming to evade trigger tokens.


Guan et al.~\cite{guan2025ufid} introduce UFID, a backdoor detection method based on the \textit{graph density score}. UFID generates multiple image variations from a perturbed input sample and constructs a fully connected graph, where each node represents a generated image and edge weights reflect pairwise image similarity. For a given input $x_i$, its similarity graph is denoted as $G_i = (V_i, \mathcal{E}_i)$, where $|V_i| = |M|$ is the set of vertices (i.e., generated images) and $\mathcal{E}_i$ is the set of weighted edges. Each edge between a pair of images $(u, v \in V)$ is assigned a similarity score $Sc(u, v)$. The core insight is that backdoor samples tend to produce highly similar outputs, resulting in a denser graph structure. The graph density $DS(G_i)$ of the similarity graph is defined as:
\begin{equation}
DS(G_i) = \frac{\sum_{m < n} Sc\left(\varepsilon(y_i^m), \varepsilon(y_i^n)\right)}{|M|(|M|-1)}
\end{equation}
where $y_i^m$ and $y_i^n$ are the $m$-th and $n$-th images in the batch generated from $x_i$, $\varepsilon(\cdot)$ is an image encoder, and $|M|$ is the total number of image pairs. A backdoor sample is detected when $\text{DS}(G_i)$ exceeds a predefined threshold $\tau$.


\begin{remark}
\label{remark_means_security_enhance}
We observe that few studies focus on developing defense mechanisms against backdoor attacks, likely due to two main challenges: \textbf{(1)} Backdoor attacks are inherently stealthy, with triggers that activate only under specific conditions. 
Designing general defenses that can reliably identify and neutralize such attacks is highly challenging and can be as complex as solving an NP-hard problem \cite{huang2022embedding}. \textbf{(2)} Effective defenses must also maintain the model's performance on benign data, making this trade-off a challenging task. \textbf{(3)} The multi-modality attack surface makes traditional defense methods for backdoor attacks difficult to apply to DMs.
\end{remark}

\noindent \textbf{Enhancement for Privacy:}\label{enhance_privacy} Most enhancement efforts for privacy focus on techniques like data deduplication, data augmentation, and differential privacy. 

Duan et al. \cite{duan2023diffusion} investigate existing methods for mitigating model overfitting, such as data augmentation (Cutout, RandomHorizontalFlip, RandAugment), differential privacy stochastic gradient descent (DP-SGD), and \(\mathcal{L}_2\) regularization, to enhance privacy. However, their experimental results show that DDPM training with these defense methods fail to converge.

Carlini et al. \cite{carlini2023extracting} employ data deduplication, using the Imagededup tool \cite{idealods2019imagededup}, to remove similar images and mitigate model memorization. They also apply DP-SGD to DM, but encounter consistent training divergence on CIFAR-10, resulting in training failure.

Wen et al. \cite{wen2024detecting} proposed memorization mitigation methods by indicating the significance score of individual tokens in relation to memorization. Given a prompt embedding \( e \) of prompt \( p \) with \( N \) tokens, they define the significance score \(\text{SS}_{e^i}\) for each token \(e\) at position \( i \in [0, N \!-\! 1] \) as:
\begin{equation}
\begin{aligned}
\text{SS}_{e^i} & = \frac{1}{T} \sum_{t=1}^{T} \| \nabla_{e_i} \mathcal{L}(z_t, e) \|^2, \\
\mathcal{L}(z_t, e) & = \| \epsilon_\theta(z_t,t \mid e) - \epsilon_\theta(z_t,t \mid {\emptyset}) \|^2,
\end{aligned}
\end{equation}
where \(\mathcal{L}\) is the training objective (classifier-free guidance) for minimization. \(z_t\) denotes the latent representation. A token with a higher significance score is more likely to be linked to memorization and can thus be rephrased or excluded before initiating a new generation.

Ren et al. \cite{ren2024unveiling} propose mitigation methods for both inference-time and training-time. During training, they remove samples from the mini-batch if their attention entropy, as defined in Eq.~\eqref{entropy_attention}, exceeds a pre-defined threshold, identifying them as memorized samples. For inference-time mitigation, they reduce the influence of trigger tokens by increasing the attention score of the first token. This is achieved by adjusting the input logits to the softmax operator in the cross-attention mechanism, based on the observation that trigger tokens (which are strongly associated with memorization) tend to exhibit high attention scores. 

Somepalli et al. \cite{somepalli2023understanding} observe that text conditioning plays a key role in data replication. To mitigate this issue, they propose several strategies for randomizing text conditioning information: (1) generating 20 captions for each image using BLIP \cite{li2023blip} and randomly sampled during training; (2) adding Gaussian noise to text embeddings; (3) randomly replacing the caption of an image with a random sequence of words; and (4) randomly selecting a word from the caption and inserting it into a random position.

Jeon et al. \cite{jeon2024understanding} introduce Sharpness-Aware Initialization for Latent Diffusion
(SAIL), an inference-time mitigation strategy that reduces memorization by selecting initial diffusion noise from regions of lower sharpness, SAIL steers the diffusion process toward smoother probability regions, mitigating memorization without retraining. This approach is based on their observation that memorization correlates with regions of high sharpness in the probability landscape (see Sec.~\ref{sec_Falsification}, ``Falsification for Privacy'').


Chen et al.~\cite{Chen_2025_CVPR} discuss the privacy-utility trade-off in mitigating memorization of T2I DMs. They propose a Prompt Re-Anchoring strategy that reduces magnitude by steering predictions away from memorized images. Rather than discarding the prompt embedding \(e_p\) when memorization is detected, they treat \(e_p\) as a valuable anchor point and re-anchor the generation process by replacing the unconditional prediction in classifier-free guidance with the \(e_p\)-conditioned prediction, thereby diverting generation away from memorization.

\begin{remark}
\label{remark_means_privacy_enhance}
    Most studies rely on traditional defense methods like DP and data augmentation. However, these commonly used defense techniques have been shown to fail in achieving the desired results in T2I DMs \cite{duan2023diffusion, carlini2023extracting}. These findings are based on empirical results and lack a theoretical foundation. While current mitigation strategies for memorization are relatively thorough and effective, they are largely built directly upon their respective detection methods. However, there remains a lack of in-depth analysis on the underlying causes of memorization, and a unified, all-in-one analytical framework is still needed to better understand this phenomenon. 
\end{remark}

\noindent \textbf{Enhancement for Fairness:} All fairness studies focus on mitigating biases in generated images from different aspects. Struppek et al. \cite{struppek2023exploiting} observe that simple homoglyph replacements in prompt can induce the model to generate culturally biased images. They propose a teacher-student procedure by fine-tuning a text encoder \(\tau_d\) (student) to minimize the embedding similarity between prompts containing homoglyphs and their Latin-only counterpart from another trained encoder \(\tau\) (teacher) by optimizing the loss function:
\begin{align}
\mathcal{L} = & \frac{1}{|B|} \sum_{x \in B} -S(\tau(x), \tau_d(x)) \nonumber \\
              & + \sum_{h \in H} \frac{1}{|B_h|} \sum_{x' \in B_h} -S(\tau(x'), \tau_d(x' \oplus h)),
\end{align}
\(S\) denotes the cosine similarity, \(B\) and \(B_h\) represent prompt batches. The operator \(\oplus\) indicates the replacement of a single predefined Latin character in a prompt \(x' \in B_h\) with its corresponding homoglyph \(h \in H\). Therefore, the first term ensures that for prompts \(x \in B\), the computed embedding of \(\tau_d\) is close to the embeddings of \(\tau\), thereby preserving the general utility of the encoder. The second term updates \(\tau_d\) to map embeddings for prompts containing homoglyph \(h \in H\) to the corresponding embedding of their Latin counterpart, ensuring invariance against certain homoglyphs. 

Zhang et al. \cite{zhang2023iti} proposed ITI-GEN, which leveraged available reference images to train a set of prompt embeddings that can represent all desired attribute categories \(m \in M\) to generate unbiased images. They designed direction alignment loss \(\mathcal{L}_{dir}^{m}\) and semantic consistency loss \(\mathcal{L}_{sem}^{m}\) to train those inclusive prompt embedding:
\begin{equation}
\begin{aligned}
    \mathcal{L}_{dir}^{m}& = 1 - (\Delta_{I}^m(i,j), \Delta_P^m(i,j)), \\
    \mathcal{L}_{sem}^{m} &= max(0, \lambda - S(\tau(t), \tau(x)),
\end{aligned}
\end{equation}
where the image direction \(\Delta_{I}\) denotes the difference between the average image embeddings of two attribute categories \(i \& j\) , while the prompt direction \(\Delta_{P}\) is the difference between their average prompt embeddings. Hence, \(\mathcal{L}_{dir}\) aims to facilitate the prompt learning of more meaningful and nuanced differences between images from different categories. \(\mathcal{L}_{sem}\) aims to prevent language drift by maximizing the cosine similarity \(S\) between the learned prompts \(t\) and the original prompt \(x\) , \(\lambda\) is a hyperparameter.

Kim et al. \cite{kim2023stereotyping} proposed a de-stereotyping framework for a fair T2I model by soft prompt tuning. They designed a de-stereotyping loss \(\mathcal{L_{DS}}\) and a regularization loss \(\mathcal{L}_{reg}\) to train the de-stereotyping prompt embedding \(e'\) while the original prompt embedding \(e\) is frozen. The symbol \(\oplus\) indicates that \(e'\) is appended before \(e\):
\begin{equation}
\begin{aligned}
    \mathcal{L_{DS}} & = \mathbb{E}_\mathcal{Y}[{cross\_entropy}(\hat{t}, t)], \\
\mathcal{L}_{reg} & = \left\| z ( \tau ( e'\oplus e) ) - z ( \tau( e^t) ) \right\|_2,
\end{aligned}
\end{equation}
\(\mathcal{L}_{DS}\) encourages the generated images to be classified under diverse attributes by a fixed zero-shot attribute classifier (e.g., CLIP). Here, \(\hat{t}\) denotes the predicted attribute of a generated image, \(t\) is the corresponding pseudo attribute label, and \(y \in \mathcal{Y}\) represents the generated image. The regularization loss \(\mathcal{L}_{reg}\) ensures semantic consistency by minimizing the distance between the latent representations of the edited prompt and an anchor prompt \(e^t\), which is constructed by explicitly inserting the attribute \(t\) into the original prompt. For example, given the text "A photo of a doctor" with a pseudo label "female", the anchor prompt becomes "A photo of a female doctor". Here, \(z(\cdot)\) denotes the latent representation conditioned on the given text.


Friedrich et al. \cite{friedrich2023fair} propose Fair Diffusion, a method that builds on biased concepts in a model and adjusts them to enhance fairness during inference. They developed several metrics to investigate sources of gender occupation bias in SD and provide a formal definition of fairness (cf.~Def.~\ref{def_fairness}). They use an image editing tool Sega \cite{brack2023sega} to mitigate bias.

Shen et al. \cite{shen2023finetuning} design a  distributional alignment loss \(\mathcal{L}_{align}\) that steers specific attributes of the generated images towards a user-defined target distribution. They define \(\mathcal{L}_{align}\) as the cross-entropy loss \(w.r.t.\) these dynamically generated targets, with a confidence threshold $C$:
\begin{equation}
    \mathcal{L}_{align} = \frac{1}{N} \sum_{i=1}^{N} I[c^{(i)} \geq C] \mathcal{L}_{\text{CE}}(h(x^{(i)}), y^{(i)}),
\end{equation}
where \(h(x^{(i)})\) is the prediction of a pre-trained classifier and \(y^{(i)}\) is the target class, \(c^{(i)}\) is the confidence of the target and \(N\) is the number of generated images. Minimizing the loss function corresponds to reducing the distance between the attributes of the generated images and the user-defined target distribution.

Zhou et al.\cite{zhou2024association} propose Mitigating association-engendered stereotypes (MAS) which consists of the Prompt-Image-Stereotype CLIP (PIS CLIP) and the Sensitive Transformer. PIS CLIP learns the relationships between prompts, images, and stereotypes to map prompts to stereotype representations, while the Sensitive Transformer constructs sensitive constraints based on these mappings. These constraints are then embedded into the T2I model to guide the stereotype probability distribution toward the stereotype-free distribution.

Kim et al.~\cite{kim2025rethinking} propose weak guidance, a debiasing method that requires no additional training. It mitigates bias by guiding the random noise toward minority regions (typically associated with bias) while preserving semantic integrity. This is achieved by injecting noise into the text condition embeddings. Specifically, for a given attribute \(k\) (e.g., ``female'' or ``male''), they define the attribute direction as \(\mathbf{a}_k = \phi(k) - \phi("")\), where \(\phi("")\) is the text embedding of an empty string. They then apply \(\mathbf{a}_k\) only to positions starting from the \texttt{[EOS]} token to get the final adjusted embedding: \(\hat{\mathbf{c}} = \mathbf{c} + \mathbf{m} \odot \mathbf{a}_k\), where \(\mathbf{m}\) is a binary mask such that \(m_i = 1\) if position \(i \geq [\text{EOS}]\), and \(0\) otherwise. During inference, \(\mathbf{c}\) and \(\hat{\mathbf{c}}\) are alternated across the denoising steps, ensuring the target attribute is gradually incorporated without disrupting the original semantics.

Li et al.~\cite{li2024self} propose a self-supervised method to discover interpretable latent directions for user-defined concepts, enabling fair and safe image generation. Given a pretrained conditional DM $\epsilon_\theta(y_t, \tau_\theta(x), t)$, where $\tau_\theta(x)$ encodes text prompt $x$, images $y^+$ are generated using prompt $x^+$ containing the concept.  
A concept vector $c \in \mathbb{R}^D$ is optimized with a modified prompt $x^-$ that omits the concept.  
The vector $c$ is injected into the U-Net bottleneck latent space and learned by minimizing reconstruction error during denoising, compensating for missing information in $\pi(x^-)$. the concept vector is found by
\begin{equation}
c^* = \arg\min_c \sum_{y,x \sim \mathcal{D}} \sum_{t \sim [0,T]} \left\| \epsilon - \epsilon_\theta(y_t^+, t, \tau_\theta(x^-), c) \right\|^2,
\end{equation}
where $y_t^+$ is the noisy image generated from $x^+$, and $\epsilon$ is the target noise. For fair generation, the optimized concept vector is uniformly sampled and added to latent activations to ensure balanced representation in generated images.


Li et al.~\cite{li2025fair} propose Fair Mapping to achieve fair generation in T2I DMs by modifying the prompt representation. They introduce a linear network $M(\cdot)$ that transforms the prompt embedding into a debiased space, enabling demographically balanced image outputs. The training objective of Fair Mapping is defined as:
\begin{align}
\mathcal{L} &= \mathcal{L}_{\text{text}} + \lambda \mathcal{L}_{\text{fair}}, \\
\mathcal{L}_{\text{text}} &= \frac{1}{|A| + 1} \left( \|v - e\|_2^2 + \sum_{a_j \in A} \|v_j - e_j\|_2^2 \right),  \\
\mathcal{L}_{\text{fair}} &= \sqrt{ \frac{1}{|A|} \sum_{a_j \in A} \left( d(v, v_j) - \overline{d(v, \cdot)} \right)^2 },
\end{align}
where $e$ and $e_j$ are the original and attribute-conditioned prompt embeddings, and $v = M(e)$, $v_j = M(e_j)$ are their transformed versions, where \(a_j \in A\) indicates sensitive attributes. $d(v,v_j)$ denotes Euclidean distance, and $\overline{d(v, \cdot)}$ is the average distance between $v$ and all $v_j$. Eq.~(39) is the total loss used to train the Fair Mapping network, balancing semantic consistency and fairness via the hyperparameter $\lambda$.  
The semantic consistency loss \(\mathcal{L}_{\text{text}}\) ensures that transformed embeddings preserve the original prompt's meaning. The fairness loss \(\mathcal{L}_{\text{fair}}\) minimizes the variance in distances to sensitive attributes, thus mitigating bias in the prompt space.



Jiang et al.~\cite{jiang2024mitigating} propose a Linguistic-aligned Attention Guidance module to mitigate social bias in multi-face image generation. They identify semantic regions \( SR_t \) associated with biased tokens \( t \) by correlating cross-attention maps \( CA_t \) with semantic clusters \( SC^i \), which are derived from self-attention using spectral clustering. The clusters \( SC^c \) represent different semantic segmentations, where \( c \) denotes the number of clusters. The semantic regions are defined as:
\begin{equation}
SR_t = \sum_{i=1}^{c} (SC^i) \cdot \mathbb{I}_\tau \left( \frac{\sum (SC^i \cdot CA_t)}{\sum (SC^i)} > \tau \right),
\end{equation}
where the agreement score \(\frac{\sum (SC^i \cdot CA_t)}{\sum (SC^i)}\) measures the relevance between tokens and image regions. This enables precise localization of bias-related content in the generated images, which in turn allows targeted debiasing operations on these regions to reduce stereotypical associations.

\begin{remark}
\label{remark_mean_fairness_enhance}
    Existing methods to enhance fairness can be classified into three types:  \textbf{(1)} Fine-tuning: Adjusting the text encoder \cite{struppek2023exploiting} or targeting image attribute distribution \cite{shen2023finetuning}. \textbf{(2)} Training Auxiliary Unbiased Text Embeddings: Incorporating unbiased attributes through additional text embeddings \cite{kim2023stereotyping, zhang2023iti, li2024self, li2025fair}. \textbf{(3)} Guiding the image generation process: Leveraging tools to modify and control the generation pipeline to ensure fairness \cite{friedrich2023fair, kim2025rethinking, zhou2024association, jiang2024mitigating}, detailed information is provided in Table~\ref{table_fairness}.
\end{remark}

\begin{table}[htbp]
\caption{Overview of fairness in T2I DMs.}
\resizebox{\linewidth}{!}{
\begin{tabular}{c|c|c|c|c}
\hline
\textbf{Paper} & \textbf{Bias Source} & \textbf{Enhancement} & \textbf{Model} & \textbf{Time} \\ \hline \rule{0pt}{10pt}
Bansal \cite{bansal2022well}  & text encoder & No & \parbox[c]{3cm}{\centering SD; DALL-Emini; minDALL-E} & 2022 \\ \hline 
Struppek \cite{struppek2023exploiting}  & text encoder; dataset & Yes & \parbox[c]{3cm}{\centering SD; DALL-E 2 } & 2023 \\ \hline
Zhang \cite{zhang2023iti}  & text encoder & Yes & \parbox[c]{3cm}{\centering SD} & 2023 \\ \hline
Friedrich \cite{friedrich2023fair}  & text encoder; dataset; U-Net & Yes & \parbox[c]{3cm}{\centering SD} & 2023 \\ \hline
Kim \cite{kim2023stereotyping}  & text encoder & Yes & \parbox[c]{3cm}{\centering SD} & 2023 \\ \hline
Bianchi \cite{bianchi2023easily}  & dataset & No & \parbox[c]{3cm}{\centering SD} & 2023 \\ \hline
Shen \cite{shen2023finetuning}  & text encoder; U-Net & Yes & \parbox[c]{3cm}{\centering SD} & 2023 \\ \hline
Luccioni \cite{luccioni2024stable}  & dataset & No & \parbox[c]{3cm}{\centering SD; DALL-E 2} & 2024 \\ \hline

Chinchure \cite{chinchure2024tibet} & dataset & Yes & \parbox[c]{3cm}{\centering SD} & 2024 \\ \hline

Zhou \cite{zhou2024association} & dataset & Yes & \parbox[c]{3cm}{\centering SD} & 2024 \\ \hline 

D’Inca \cite{d2024openbias} & dataset & No & \parbox[c]{3cm}{\centering SD} & 2024 \\ \hline

Li \cite{li2024self} & -- & Yes & \parbox[c]{3cm}{\centering SD} & 2024 \\ \hline

Jiang ~\cite{jiang2024mitigating} & -- & Yes & \parbox[c]{3cm}{\centering SD} & 2024 \\ \hline

Kim \cite{kim2025rethinking} & -- & Yes & \parbox[c]{3cm}{\centering SD } & 2025 \\ \hline

Li \cite{li2025fair} & -- & Yes & \parbox[c]{3cm}{\centering SD} & 2025 \\ \hline

Huang \cite{huang2025implicit} & -- & Yes & \parbox[c]{3cm}{\centering SD} & 2025 \\ \hline

\end{tabular}
}
\label{table_fairness}
\end{table}

\noindent \textbf{Enhancement for Explainability:}
Hertz et al. propose Prompt-to-Prompt \cite{hertz2023prompttoprompt}, an image editing framework that controls the relationship between the spatial layout of the image and each word in the prompt through cross-attention layers (cf. Eq.~\eqref{U-Net_cross_attention}). By visualizing cross-attention maps in U-Net, this method allows for the observation of more complex visual interactions, providing a clearer interpretation of the internal workings of the text guidance function during the generation process.

Similarly, Tang et al. \cite{tang2023daam} propose DAAM, a T2I attribution analysis method for SD. They design pixel-level attribution maps by up-scaling and aggregating cross-attention word–pixel scores in the denoising sub-network (U-Net) to interpret the generation process. DAAM is also applied to the semantic segmentation task to evaluate its accuracy.

Lee et al. \cite{lee2023diffusion} propose Diffusion Explainer, an interactive tool designed for non-experts that provides a visual overview of each component of SD. It compares how image representations evolve over refinement timesteps when guided by two related text prompts, highlighting how keyword differences in the prompts affect the evolution trajectories starting from the same initial random noise. The main objective is to visualize how keywords in the text prompt affect image generation.

Evirgen et al. \cite{evirgen2024text} introduce four methods to help understand the generation process. 1) The Keyword Heat Map method uses cross-attention maps to highlight pixel regions most influenced by specific keywords. 2) The Redacted Prompt Explanation technique leverages CLIP to measure similarity between original and modified images (generated by randomly removing a set of keyword from original prompt). 3) Keyword Linear Regression approximates the image generation process as a linear combination of keywords, representing their contributions as linear weights. 4) The Keyword Image Gallery aims to create a tailored collection of images for each keyword, highlighting the keyword’s influence on the image generation process. All these methods are designed to interpret how specific keywords affect both the generation process and the resulting images.

Chefer et al.~\cite{chefer2024the} propose CONCEPTOR to interpret how a T2I DM represents a textual concept by decomposing it into a small set of interpretable elements. Given a concept prompt \(x^c\), CONCEPTOR learns a pseudo-token \(w^*\) as a weighted combination of a subset of vocabulary tokens:
\begin{align}
w^* &= \sum_{i=1}^{n} \alpha_i w_i, \quad \text{s.t. } w_i \in V,\ \alpha_i \geq 0, \notag \\
\mathcal{L} &= \mathcal{L}_{\text{rec}} + \lambda_{\text{sparsity}} \cdot (1 - \cos(w^*, w^*_N)), \label{eq:conceptor}
\end{align}
where \(V\) is the vocabulary, \(n \ll N\), and \(w^*_N = \sum_{i=1}^{N} f(w_i) w_i\) is the full vocabulary-based representation obtained from a learned 2-layer MLP \(f(w)\). The objective \(\mathcal{L}\) combines a reconstruction loss with a sparsity regularization to ensure that \(w^*\) remains interpretable while approximating the dense representation \(w^*_N\) from the entire vocabulary.

\begin{remark}\label{remark_means_explain}
    As shown in Table~\ref{table_explain}, while attention heatmaps have been used to clarify the image generation process in T2I DMs for local inputs \cite{evirgen2024text, tang2023daam} and an interactive framework \cite{lee2023diffusion} has been explored, many other types of XAI methods commonly used in traditional DL tasks have not yet been studied for T2I DMs, e.g., global and perturbation-based XAI. 
    Furthermore, it is well-known that XAI methods themselves are unrobust yielding wrong explanations when subject to small input perturbations \cite{ghorbani2019interpretation,zhao_baylime_2021,Huang_2023_ICCV}. Therefore, the robustness of XAI in T2I tasks requires further investigation.
\end{remark}

\begin{table}[htbp]
\caption{Overview of explainability in T2I DMs.}
\resizebox{\linewidth}{!}{
\begin{tabular}{c|c|c|c}
\hline
\textbf{Paper} & \textbf{Explainer} & \textbf{Model} & \textbf{Time} \\ \hline
Hertz \cite{hertz2023prompttoprompt} & attention map &  Imagen & 2023 \\ \hline
Lee \cite{lee2023diffusion}  & interactive visualization tool &  SD & 2023 \\ \hline
Tang \cite{tang2023daam} & attention map & SD & 2023 \\ \hline
Evirgen \cite{evirgen2024text}  & keyword information & SD & 2024 \\ \hline 

Chefer \cite{chefer2024the}  & Concept interpretability & SD & 2024 \\

\hline
\end{tabular}
}
\label{table_explain}
\end{table}

\noindent \textbf{Enhancement for Factuality:}
Lim et al. \cite{lim2024addressing} investigate factuality issues in T2I DMs by categorizing types of hallucination as defined in Def.~\ref{def_Factuality}. They propose incorporating factual images from external sources (images retrieved by Google’s Custom Search JSON API) to enhance the realism of generated images, analogous to how external knowledge is incorporated in LLMs \cite{thoppilan2022lamda}.

Additionally, controllable image generation aims to mitigate hallucinations through \textit{pre-processing} methods, which use textual conditions to guide the image generation process. Cao et al. \cite{cao2024controllable} provide a comprehensive survey of controllable T2I DMs. Zhang et al. \cite{zhang2023adding} introduce ControlNet, which adds spatial control to pre-trained T2I models using ``zero convolution'' links, enabling stable training and flexible image control. Mou et al. \cite{mou2024t2i} develop lightweight adapters that provide precise control over image color and structure, trained independently from the base T2I models.

Furthermore, image editing can reduce hallucinations through \textit{post-processing} techniques. This approach involves altering an image's appearance, structure, or content \cite{hertz2023prompttoprompt, mo2024dynamic}. Zhang et al. \cite{zhang2023sine} propose a flexible method using natural language, combining model-based guidance with patch-based fine-tuning to enable style changes, content additions, and object manipulations. Kim et al. \cite{kim2022diffusionclip} introduce DiffusionCLIP, a robust CLIP-guided method for text-driven image manipulation. For more work on image editing, refer to discussions on model-agnostic applications \ref{Domain-agnostic}.

\begin{remark}
\label{remark_enhance_factuality}
Current approaches to improving factuality in T2I DMs often adapt methods from LLMs \cite{lim2024addressing, su-etal-2022-read} or auxiliary techniques, such as controllable strategies and image editing. However, specific studies on the causes and extent of hallucinations unique to T2I DMs are significantly underexplored compared to research in other fields like LLMs.
\end{remark}


\subsection{Benchmarks and Applications}

\subsubsection{Benchmarks}
Recent progress in T2I DMs have led to the creation of a variety of benchmarks aimed at systematically evaluating their performance and accuracy, as summarized in Table~\ref{table_benchmarks_applications}. These benchmarks primarily assess the functional capabilities of T2I models, such as image quality, coherence between text prompts and generated images \cite{otani2023toward, park2021benchmark, huang2023t2i, sun2024journeydb, baiqi-li2024genaibench, yu2022scaling}. For example, Imagen \cite{saharia2022photorealistic} introduced \textbf{DrawBench} to evaluate T2I models across various dimensions like compositions, conflicts, and writing, alongside image quality. \textbf{HE-T2I} \cite{petsiuk2022human} suggests 32 possible aspects for benchmarking T2I models, but focuses on just three: counting, shapes, and faces. \textbf{TISE} \cite{dinh2022tise} provides a bag of metrics for evaluating models based on positional alignment, counting, and fidelity. \textbf{T2I-CompBench++} \cite{huang2025t2i} focuses on evaluating the compositional abilities of T2I models by introducing 8,000 challenging prompts spanning attributes, object relationships, numeracy, and spatial reasoning. 

Beyond functional evaluation, some benchmarks target non-functional properties such as fairness and factuality. For instance, \textbf{DALL-EVAL} \cite{Cho2023DallEval} assesses three core visual reasoning skills—object recognition, object counting, and spatial relation understanding—while also considering social bias in terms of gender and race. \textbf{HRS-Bench} \cite{bakr2023hrs} measures 13 skills across five categories—accuracy, robustness, generalization, fairness, and bias—covering 50 scenarios including fashion, animals, transportation, food, and clothes. \textbf{ENTIGEN} \cite{bansal2022well} covers prompts to evaluate bias across three axes: gender, skin color, and culture. It is designed to study changes in the perceived societal bias of T2I DMS when ethical interventions are applied.  \textbf{T2I-FactualBench} \cite{huang2024t2i} focuse on knowledge-intensive image generation. It introduces a three-tiered benchmark to assess factual understanding, ranging from basic factual recall to complex knowledge composition, and employs a multi-round VQA-based evaluation framework. \textbf{DoFaiR}~\cite{wan-etal-2024-factuality} examines the tension between diversity prompting and demographic factuality in T2I models, especially for historical figures. It introduces an evaluation pipeline to measure this trade-off and proposes Fact-Augmented Intervention (FAI) to enhance demographic accuracy without sacrificing diversity. \textbf{HEIM} \cite{lee2023holistic} introduces a benchmark covering 12 evaluation aspects—including alignment, image quality, aesthetics, originality, reasoning, knowledge, bias, toxicity, fairness, robustness, multilinguality, and efficiency—across 62 curated scenarios and 26 SOTA T2I DMs. T2ISafety \cite{li2025t2isafety} focuses on three dimensions: toxicity, fairness, and bias. It defines a hierarchy of 12 tasks and 44 categories, supported by 70K prompts and 68K annotated images, to evaluate and detect critical safety risks across 15 diffusion models.

\begin{remark}
\label{remark_benchmark}
    Existing benchmarks primarily focus on performance and accuracy, emphasizing the core functionality of T2I models. While some benchmarks have been developed to assess non-functional properties, most of them concentrate narrowly on fairness and factuality. Other important trustworthiness dimensions discussed in our survey—such as explainability, security, robustness, and privacy—remain relatively underexplored and lack systematic evaluation frameworks.
\end{remark}

\subsubsection{Applications}
Recent advancements in T2I DMs have spark interest in various compelling applications across specific domains such as Intelligent Vehicles, Healthcare, and a series of Domain-agnostic Generation Tasks, as shown in Table.~\ref{table_benchmarks_applications}. While DMs, or more generally large foundation models, are finding broader applications in fields such as robotics, material design, and manufacturing, these applications are not specifically related to T2I tasks and are therefore beyond the scope of this survey.

\textbf{T2I DMs for Intelligent Vehicle} T2I DMs are used in Intelligent Vehicle for safety-critical scenarios generation \cite{guo2023controllable, gannamaneni2024exploiting, cheng2024instance}, and open-vocabulary panoptic segmentation \cite{xu2023open}, e.g., Gannamaneni et al. \cite{gannamaneni2024exploiting} propose a pipeline to generate augmented safety-critical scenes from the Cityscapes dataset using SD and OpenPose-based ControlNet.

\textbf{T2I DMs for Healthcare}  T2I DMs have been applied to various medical downstream tasks, particularly medical image synthesis \cite{kazerouni2023diffusion, cho2024medisyn,sagers2023augmenting,chambon2022roentgen, XuMedSyn2024, JangTauPETGen2023}. For example, Sagers et al. \cite{sagers2023augmenting} employ DALL-E to synthesize skin lesions across all Fitzpatrick skin types. Xu et al. \cite{XuMedSyn2024} generate high-quality 3D lung CT images guided by textual information based on GAN and DM. Similarly, Jang et al. \cite{JangTauPETGen2023} used SD to produce realistic tau PET and MR images of subjects.

\textbf{T2I DMs for Domain-agnostic Generation Tasks} \label{Domain-agnostic} Beyond specific domain applications, T2I DMs have been widely used in various domain-agnostic generation tasks, including general image editing \cite{kim2022diffusionclip, chandramouli2022ldedit, hertz2023prompttoprompt, zhang2023sine}, 3D generation \cite{hollein2024viewdiff, poole2023dreamfusion, liu2023zero, liu2024one}, and video generation \cite{wu2023tune, khachatryan2023text2video, ho2022video, esser2023structure, hong2023cogvideo}. E.g., Kim et al. \cite{kim2022diffusionclip} propose DiffusionCLIP for both in-domain and out-of-domain text-driven image manipulation. DreamFusion \cite{poole2023dreamfusion} optimize a 3D representation through score distillation sampling from T2I DMs. Text2Video-Zero \cite{khachatryan2023text2video} leverage SD to achieve zero-shot text-to-video generation.
Additionally, a series of works adopt T2I DMs to generate adversarial examples \cite{truong2025attacks, chen2023content, 10716799}, leveraging their ability to produce semantically rich and photorealistic images to enable more natural and transferable attacks compared to traditional pixel-level perturbation methods. For example, Chen et al.~\cite{chen2023content} propose a content-based unrestricted adversarial attack that leverages DMs as natural image manifolds to generate photorealistic and transferable adversarial examples by optimizing in the latent space. In a separate work, Chen et al.~\cite{10716799} introduce DiffAttack, which crafts perturbations in the latent space of diffusion models rather than directly manipulating pixel values, resulting in imperceptible yet semantically meaningful adversarial examples.

\begin{remark}
\label{remark_application}
    Whether in specific fields such as intelligent vehicles and healthcare care or in domain-agnostic generation tasks, the focus has solely been to achieve high performance and precision, with little attention to ensure that models are trustworthy for real-world applications.
\end{remark}

\section{Discussion}
Based on the in-depth analysis of the six non-functional properties around trustworthiness and four means discussed in the aforementioned papers, we have summarised several key findings that could guide future research on trustworthy T2I DMs.

\textbf{Limitation and Direction for Robustness:}
As per Remarks~\ref{remark_definition_robust}, \ref{remark_means_robust_falsification}, \ref{remark_means_robust_enhance}, \ref{remark_means_robust_VV}:
\textbf{(1)} Existing research on T2I DMs primarily focuses on worst-case (maximum loss) robustness, leaving binary and maximum radius robustness (cf.~Fig.~\ref{fig_types_robustness}) largely unexplored. This is partially due to the difficulty in defining a small norm-ball region for text inputs with the same semantic meaning, as quantifying semantic distance in the text space remains a challenging task in the field of natural language processing. \textbf{(2)} Almost all existing work using falsification methods has focused on the text encoder, with a notable lack of studies examining the vulnerabilities of the diffusion component. \textbf{(3)} Existing work on enhancement mainly depends on external auxiliary spellcheckers \cite{zhang2024protip, liu2023riatig} to defend against perturbations. However, there is a lack of research on improving the model's inherent robustness, such as adopting adversarial training to withstand attacks. \textbf{(4)} There have been limited efforts in V\&V studies. Only Zhang et al. \cite{zhang2024protip} proposed a verification framework, as defining a specification for T2I verification is a challenging task. Based on these key findings, several possible research directions for improving the robustness of T2I DMs include: \textbf{(1)} Developing effective metrics to quantify semantic distance in text, which would enable binary and maximum radius robustness. \textbf{(2)} Designing new attack objectives specifically targeting the diffusion process. \textbf{(3)} Exploring enhancement methods, such as adversarial training, to strengthen the model's internal resilience against adversarial perturbations.

\textbf{Limitation and Direction for Fairness}
As per Remarks \ref{remark_definition_fairness}, \ref{remark_mean_fairness_enhance} and \ref{remark_means_fairness_assess}: \textbf{(1)} Existing fairness work for T2I DMs only consider the use cases based on ``one-off'' queries \cite{friedrich2023fair}, neglecting the more complex interaction patterns that occur when users engage with AI systems over time. \textbf{(2)} Existing methods to enhance fairness can be classified into three types: Adjusting the text encoder \cite{struppek2023exploiting} or targeting image attribute distribution \cite{shen2023finetuning}; training auxiliary unbiased text embeddings \cite{kim2023stereotyping, zhang2023iti}; using image editing tools to control the images generation process \cite{friedrich2023fair}. Therefore, future research directions may include: \textbf{(1)} More formal and comprehensive fairness definitions are needed for T2I tasks, especially for interactive use cases. \textbf{(2)} New enhancement methods are required to correspond to the fairness challenges posed by interactive use cases.


\textbf{Limitation and Direction for Security:}
Limitations are observed in security studies from Remarks \ref{remark_definition_security}, \ref{remark_means_security_falsification}, \ref{remark_means_security_enhance}: \textbf{(1)} Existing backdoor works focus on static triggers, which are inflexible and easily detectable. 
\textbf{(2)} Only three studies, \cite{Wang2024T2IShield, chew2024defending, guan2025ufid}, explore mitigation and detection strategies. This challenge arises from the difficulty of detecting subtle triggers, as developing general defenses to identify and mitigate these attacks is complex and may be as difficult as solving an NP-hard problem \cite{huang2022embedding}. \textbf{(3)} Balancing the effectiveness of these defenses while preserving the model's performance adds further complexity. \textbf{(4)} No study has explored structure-modified backdoors, where hidden backdoors are added by altering the model’s structure, as seen in traditional DL systems. Therefore, these limitations highlight potential research directions: \textbf{(1)} Investigating dynamic backdoor triggers, which are generated by specific systems and display random patterns and locations. \textbf{(2)} Developing adaptive defense frameworks that change based on the nature of the trigger. \textbf{(3)} Exploring the trade-off between performance and security \cite{li2020invisible, borgnia2021strong}, which presents another promising research avenue. \textbf{(4)} Studying structure-modified backdoors to uncover insights for advancing model security.

\textbf{Limitation and Direction for Privacy:}
As per Remarks \ref{remark_def_privacy} and \ref{remark_means_privacy_enhance}: \textbf{(1)} Privacy studies for T2I DMs mainly focus on memorization, MIAs and data extraction attacks. This focus stems from the added complexity of T2I DMs, as their multi-modal nature presents more challenges than single-modal systems. Additionally, the intricate training objectives and underlying algorithms of T2I DMs have led to less exploration of model extraction and property inference attacks. \textbf{(2)} Existing privacy enhencement methods, like DP and data augmentation, often fall short for T2I DMs \cite{duan2023diffusion, carlini2021extracting}. Therefore, further research can be conducted: \textbf{(1)} Investigating attacks specifically targeting the T2I DMs themselves, such as model extraction, which can reveal insights into the specific data sampling and generation algorithms used. \textbf{(2)} Conducting \textit{theoretical} studies to understand why commonly used privacy protection methods fail for T2I DMs, in addition to empirical evidence. This understanding will aid in designing more effective privacy enhancement strategies tailored to T2I DMs. \textbf{(3)} Although current mitigation strategies for memorization are fairly effective, they mainly rely on specific detection methods. However, there is still a lack of in-depth analysis of the underlying causes of memorization, and a unified, comprehensive analytical framework is needed to better understand this phenomenon.

\textbf{Limitation and Direction for Explainability:}
Limitations on explainability (see Remarks \ref{remark_def_explainable} and \ref{remark_means_explain}) include: \textbf{(1)} Current XAI work mainly focuses on local XAI, interpreting individual samples. \textbf{(2)} Traditional XAI methods, such as attention heatmaps, are applied to T2I DMs, and some research has explored interactive modes to visualize the image generation process and improve image quality. These findings indicate future research directions: \textbf{(1)} Adapt and apply additional XAI methods from traditional DL tasks to T2I DMs to further enhance interpretability. \textbf{(2)} Further exploration is needed to evaluate and ensure the robustness of these XAI methods \cite{zhao_baylime_2021,Huang_2023_ICCV}.

\textbf{Limitation and Direction for Factuality:}
Based on the review of factuality studies and Remarks \ref{def_Factuality} and \ref{remark_enhance_factuality}, we found: \textbf{(1)} Existing enhancement works primarily adopt methods from LLMs and auxiliary techniques (controllable strategies, image editing) to mitigate hallucinations. However, there is a lack of in-depth analysis on how and why hallucinations occur in T2I tasks. Therefore, future research should focus on: \textbf{(1)} Conducting formal assessments of hallucination phenomena in T2I DMs, including establishing formal definitions and quantification metrics to understand why and how hallucinations occur. \textbf{(2)} Designing specific methods based on these definitions and assessments to improve factuality.

\textbf{Limitation and Direction for Benchmarks and Applications:}
We summarize from Remarks \ref{remark_benchmark} and \ref{remark_application} regarding benchmarks and applications: \textbf{(1)} Existing benchmarks mainly focus on functional properties, with only limited attention to non-functional aspects such as fairness. \textbf{(2)} Both domain-agnostic and specific applications largely concern performance and accuracy, often overlooking trustworthiness. Therefore, future research directions include: \textbf{(1)} The need for comprehensive benchmarks that evaluate both functional and non-functional properties, as discussed in this paper. \textbf{(2)} Real-world applications, whether domain-specific or domain-agnostic, should ensure trustworthiness by considering non-functional properties.

We also found a common characteristic across all properties: \textbf{(1)} SD is the most frequently studied model, as it is the only open-source T2I DM. While some studies experiment with other T2I DMs like DALL-E 2 or Imagen, these are typically case studies. The adaptability and generalizability of these aforementioned trustworthy research findings to other models need further exploration.

\section{Conclusion}
This paper presents an in-depth examination of T2I DMs, offering a concise taxonomy centered on non-functional properties related to trustworthiness, highlighting the challenges and complexities within this field. We have outlined clear definitions of six key trustworthiness \textit{properties} in T2I DMs: robustness, privacy, security, fairness, explainability, and factuality. Our analysis of four primary \textit{means} — falsification, verification \& validation, assessment, and enhancement. We have showcased the solutions proposed across various studies to address these critical ethical concerns. Additionally, we also cover existing benchmarks and applications of T2I DMs, identifying key gaps and suggesting future research directions to foster more trustworthy T2I DMs. This work serves as a foundational resource for future research and development, aiming to improve the trustworthiness of T2I DMs.

\ifCLASSOPTIONcompsoc
  \section*{Acknowledgments}
\else
  \section*{Acknowledgment}
\fi

This work is supported by the British Academy's Pump Priming Collaboration between UK and EU partners 2024 Programme [PPHE24/100115], UKRI Future Leaders Fellowship [MR/S035176/1], UK Department of Transport [TETI0042], and Transport Canada [T8080-220112]. The authors thank the WMG centre of HVM Catapult for providing the necessary infrastructure for conducting this study. XH's contribution is supported by the UK EPSRC through End-to-End Conceptual Guarding of Neural Architectures [EP/T026995/1]. YZ's contribution is supported by China Scholarship Council.

\ifCLASSOPTIONcaptionsoff
  \newpage
\fi

\bibliographystyle{IEEEtran}
\bibliography{ref}
\end{document}